\documentclass[10pt,twocolumn,letterpaper]{article}

\usepackage[pagenumbers]{iccv} 

\usepackage{microtype}
\usepackage{enumitem}
\usepackage{bbding}
\usepackage{wrapfig}
\usepackage{multirow}
\usepackage{xspace}
\usepackage{colortbl}
\usepackage{pifont}
\usepackage{lipsum}
\usepackage{algorithm}
\usepackage[noend]{algpseudocode}
\usepackage{multirow}

\usepackage{color}
\definecolor{turquoise}{cmyk}{0.65,0,0.1,0.3}
\definecolor{purple}{rgb}{0.65,0,0.65}
\definecolor{darkgreen}{rgb}{0, 0.5, 0}
\definecolor{orange}{rgb}{0.8, 0.6, 0.2}
\definecolor{red}{rgb}{0.8, 0.2, 0.2}
\definecolor{darkred}{rgb}{0.6, 0.1, 0.05}
\definecolor{blueish}{rgb}{0.0, 0.3, .6}
\definecolor{light_gray}{rgb}{0.7, 0.7, .7}
\definecolor{pink}{rgb}{1, 0, 1}
\definecolor{greyblue}{rgb}{0.25, 0.25, 1}
\definecolor{tbl_no1}{rgb}{0.42, 0.76, 0.21}
\definecolor{tbl_no2}{rgb}{0.68, 0.87, 0.51}

\renewcommand{\paragraph}[1]{\vspace{.3em}\noindent\textbf{#1}}

\newcommand\blfootnote[1]{%
  \begingroup
  \renewcommand\thefootnote{}\footnote{#1}%
  \addtocounter{footnote}{-1}%
  \endgroup
}

\newcommand{\Fig}[1]{Fig.~\ref{fig:#1}}

\newcommand{\Tab}[1]{Tab.~\ref{tbl:#1}}

\newcommand{\Sec}[1]{Sec.~\ref{sec:#1}}

\makeatletter
\DeclareRobustCommand\onedot{\futurelet\@let@token\@onedot}
\def\@onedot{\ifx\@let@token.\else.\null\fi\xspace}
\def\eg{\emph{e.g}\onedot} 
\def\ie{\emph{i.e}\onedot}

\makeatother

\newcommand{\cmark}{\ding{52}}
\newcommand{\greencmark}{\color{darkgreen}{\cmark}}
\newcommand{\xmark}{\ding{56}}
\newcommand{\redxmark}{\color{darkred}{\xmark}}

\definecolor{iccvblue}{rgb}{0.21,0.49,0.74}
\usepackage[pagebackref,breaklinks,colorlinks,allcolors=iccvblue]{hyperref}

\title{Visual Chronicles:\\
Using Multimodal LLMs to Analyze Massive Collections of Images}

\author{
Boyang Deng${^1}$
$\qquad$
Songyou Peng${^{2,*}}$
$\qquad$
Kyle Genova${^{2,*}}$
\\
Gordon Wetzstein${^1}$
$\qquad$
Noah Snavely${^2}$
$\qquad$
Leonidas Guibas${^{1,2}}$
$\qquad$
Thomas Funkhouser${^2}$\\[4pt]
$^1$Stanford University
$\qquad$
$^2$Google DeepMind
}

\begin{document}

\setlength{\floatsep}{0.5\floatsep}
\setlength{\dblfloatsep}{0.5\dblfloatsep}
\setlength{\textfloatsep}{0.5\textfloatsep}
\setlength{\dbltextfloatsep}{0.5\dbltextfloatsep}
\raggedbottom

\twocolumn[{%
\renewcommand\twocolumn[1][]{#1}%
\maketitle
\begin{center}
    \captionsetup{type=figure}
    \vspace{-0.5cm}
    \includegraphics[width=.98\textwidth]{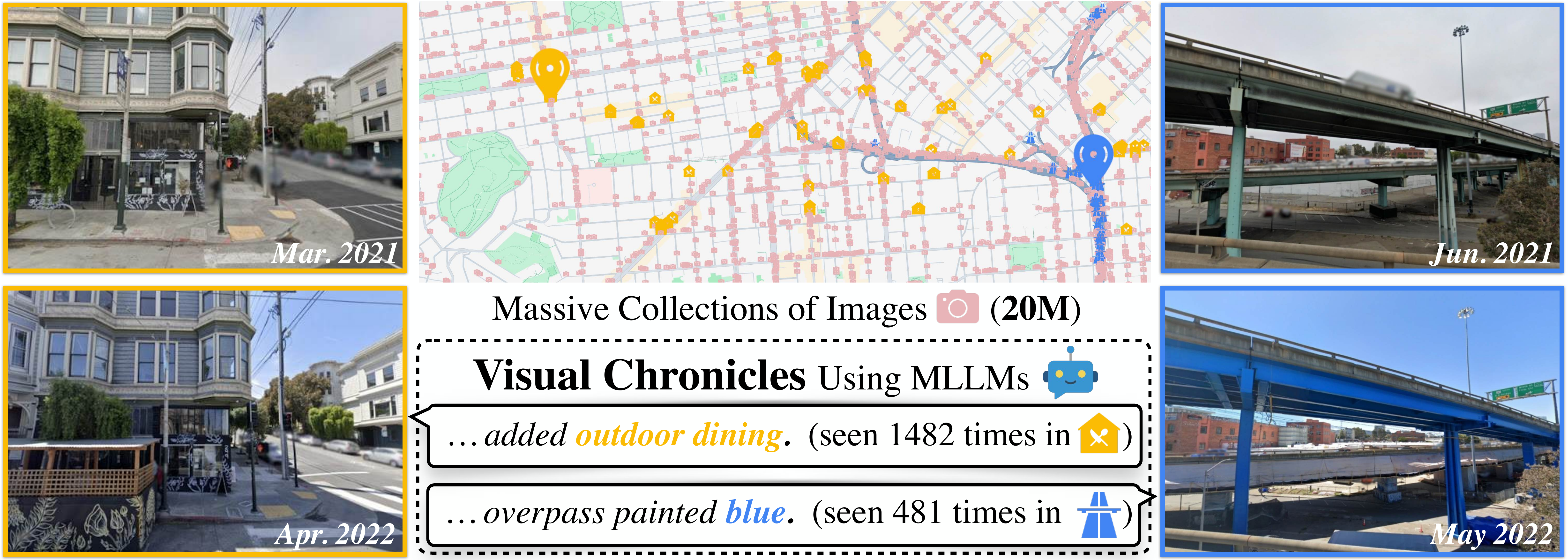} \\
    \captionof{figure}{
    \textbf{Visual Trend Discovery from Massive-Scale Collections of Images}.
    We introduce \emph{Visual Chronicles}, a novel approach that uses Multimodal LLMs (MLLMs) to make discoveries from massive collections of images over time. 
    We analyze millions of images collected over more than 10 years by Google Street View.
    As seen above, our method discovers both expected and surprising \textit{visual trends} in San Francisco---e.g. outdoor dining areas were added to many storefronts, and miles of overpass was painted blue, both with visual evidence.
    }
    \label{fig:teaser}
\end{center}
}]

\maketitle
\begin{abstract}
\label{sec:abstract}
We present a system using Multimodal LLMs (MLLMs) to analyze a large database with tens of millions of images captured at different times, with the aim of discovering patterns in temporal changes.
Specifically, we aim to capture frequent co-occurring changes (``trends'') across a city over a certain period.
Unlike previous visual analyses, our analysis answers
open-ended queries (e.g., ``what are the frequent types of changes in the city?'')
without any 
predetermined target subjects
or training labels.
These properties cast prior learning-based or unsupervised visual analysis tools unsuitable.
We identify MLLMs as a novel tool for their open-ended semantic understanding capabilities.
Yet, our datasets are four orders of magnitude too large for an MLLM to injest as context.
So we introduce a bottom-up procedure that decomposes the massive visual analysis problem into more tractable sub-problems.
We carefully design MLLM-based solutions to each sub-problem.
During experiments and ablation studies with our system, we find it significantly outperforms baselines and is able to discover interesting trends from images captured in large cities (e.g., ``addition of outdoor dining,'', ``overpass was painted blue,'' etc.).
See more results and interactive demos at  \href{https://boyangdeng.com/visual-chronicles}{our project page}.
\blfootnote{* Equal contributions. Ordered decided by a random number generator.}
\end{abstract}
\section{Introduction}
\label{sec:intro}
Searching for semantic patterns and correlations in image databases is a long-standing problem in computer vision with applications in
urban planning,
visual analytics, and other fields.
Classical research on this problem
includes
tracking changes in images within specific domains over time (\eg, ``A Century of Portraits''~\cite{ginosar2015century})
and
finding salient patterns of objects in images (\eg, estimating demographics from vehicle sightings~\cite{gebru2017using}).
However, these methods have only been demonstrated for \emph{predetermined specific} target subjects (\eg, faces~\cite{ginosar2015century} or cars~\cite{gebru2017using});
and often require \emph{labeled training} data (\eg, labeled car attributes~\cite{gebru2017using}).

In this paper, we study how to make open-ended discoveries about frequently occurring changes (``trends'') across a city within a given time period, using visual analysis of a massive collection of images without required labeled training data.
We provide a study using Google Street View as a data source~\cite{anguelov2010google}. Street View provides tens of millions of posed and timestamped images covering large cities with repeated captures spanning a decade.
Our goal is to answer open-vocabulary queries about how a city has changed over time.
In contrast to prior work, our queries are open-ended, 
and not limited to a predetermined list of subject categories.
Additionally, the data doesn't come with labeled answers to our queries, and manual labeling at scale is infeasible because the nature of the queries are unknown \emph{a priori}.

The new properties of our analysis make it a difficult problem.
Any pretrained subject-specific recognition model, \eg a car detector, is insufficient for open-ended queries.
Meanwhile, the lack of labels prevents us from training a tailored learning-based model as in prior work~\cite{gebru2017using}.
Training-free image analysis techniques, from image features~\cite{doersch2015makes} to unsupervised remote sensing~\cite{melgani2002unsupervised}, have also been used in prior work.
However, 
our goal
is to find semantically interesting insights.
This demands a holistic semantic understanding of images, \eg,  understanding that a shop is closed from visual evidence, which these prior techniques fail to offer.

Therefore, we develop a new, scalable, training-free solution to open-ended visual analysis, leveraging the considerable scene understanding capabilities of Multimodal Large Language Models (MLLMs).
Our solution is motivated by a key observation---readymade MLLMs are very effective at open-ended semantic analyses of a small collection of images.
They can detect semantically salient temporal changes with high accuracy in 20-40 images.
However, a straightforward application of MLLMs does not solve our problem due to the massive scale of the data.
For instance, simply asking an MLLM to ``name the trends'' without using images would fail, because we empirically find they mainly produce abstract answers such as economical growth.
Alternatively, we could feed all images as context to an MLLM, and ask it to ``find the trends''.
However, this does not work due to the 
inability of MLLMs to handle massive context.
Thus, we design a novel bottom-up hierarchical system to decompose the problem into smaller subtasks: detecting and summarizing local visual changes followed by finding global trends among these local changes.
For each subtask, we propose an effective MLLM-based solution.
To the best of our knowledge, our system marks the first use of MLLMs for the analysis of a database of more than a million images.

Experiments with our proposed system reveal several interesting insights.
According to results of tests with manually labeled data produced for this paper, we find that our system discovers visual trends more accurately than prior solutions or trivial use of MLLMs.
In experiments with 20 million Street View images from each of two major cities (NYC and SF), it finds several interesting trends that would be difficult to discover without detailed visual analyses (see \Fig{teaser}).

Overall, our research contributions are as follows:
\begin{itemize}
    \item We study a novel \emph{open-ended} visual analysis problem on a \emph{massive-scale} dataset \emph{without labels}.
    \item We identify MLLMs as the critical tool for such unprecedented analyses, and design a novel system for massive-scale analyses using MLLMs, bypassing the context and efficiency limitations.
    \item We present experiments using our system in two major cities yielding unexpected and informative findings.
\end{itemize}
These contributions together signal a new paradigm of computer vision studies---advancing from previous task-specific supervised training or supervised fine-tuning, to carefully designed use of MLLMs in a system like ours, for complicated large-scale tasks without labels.
\section{Related Work}
\label{sec:related_works}
\subsection{Large-Scale Image Analysis}

\paragraph{General Analysis.}
Analyzing large-scale visual data--such as datasets composed of millions of images--can unveil important characteristics of the data that cannot be observed from just a few examples.
Prior work has 
applied such analytical study to various visual data collections, leading to interesting discoveries, including distinctive visual styles~\cite{doersch2015makes, siglidis2024diffmining, karayev2013recognizing},
recurring visual patterns in art~\cite{shen2019discovering, isola2011understanding}, dataset biases and irregularities~\cite{torralba2011unbiased, boiman2007detecting, khosla2012undoing}, and socio-economic insights such as economic outcomes~\cite{jean2016combining}, urban neighborhood perceptions~\cite{dubey2016deep}, and demographic compositions~\cite{gebru2017using}.
Among these works, imagery linked to GIS~\cite{anguelov2010google} has been a popular data source~\cite{doersch2015makes, StreetStyle2017, jean2016combining, dubey2016deep, gebru2017using}, closely related to the data in our study.
While these methods have made significant contributions, they largely rely on image-driven analysis,
e.g., clustering of visual elements~\cite{doersch2015makes, StreetStyle2017}, interconnecting images through shared visual structure~\cite{heath2010image}, learning visual prototypes~\cite{van2023prototype}, average images~\cite{zhu2014averageexplorer, ginosar2015century}, color diagrams~\cite{torralba2011unbiased}, image recognition models~\cite{dubey2016deep, jean2016combining, gebru2017using}, or analysis-by-synthesis approaches with image generation models~\cite{dalens2019bilinear, chen2023s, deng2024streetscapes, siglidis2024diffmining}.
These methods primarily analyze visual information without considering temporal change.

\paragraph{Temporal Analysis.}
Building upon the foundation of large-scale image analysis, other 
methods 
examine visual data over time to uncover changes, trends, and evolving patterns~\cite{isola2011makes,martin2015time,martin20153d,naik2017computer,naik2014streetscore,ginosar2015century,zhu2014averageexplorer,laffont2014transient,xie2016transfer, StreetStyle2017}.
For example, Martin-Brualla et al.~\cite{martin2015time,martin20153d} mined time-lapse videos from internet photos to visualize landmark transformations, revealing patterns of construction, decay, and seasonal variations.
Naik et al.~\cite{naik2017computer} predicted urban change by analyzing sequential street view imagery, identifying factors that influence physical transformations in cities.
Ginosar et al.~\cite{ginosar2015century} analyzed a century of American high school yearbook photos to study the evolution of facial expressions and fashion trends, demonstrating how temporal analysis can provide sociological insights.
These methods either require specific, predetermined target categories~\cite{ginosar2015century} or labelled training data~\cite{naik2017computer},
whereas our method leverages MLLMs for open-ended analyses without the need for labeled data.
Our work can capture temporal and contextual changes more effectively, defining a new direction for future image analysis.
Note that a separate line of research focuses on analyzing video data, targeting much smaller time scales and relying on the continuity between frames~\cite{laptev2005space,zelnik2001event,lindeberg1997automatic,laptev2003interest,tran2018closer,karpathy2014large,boiman2007detecting}. In contrast, we focus on temporal analysis of non-continuous temporal image collections.

\begin{figure}[t]
\centering
\includegraphics[trim={0 0 0 3cm},clip,width=0.95\linewidth]
{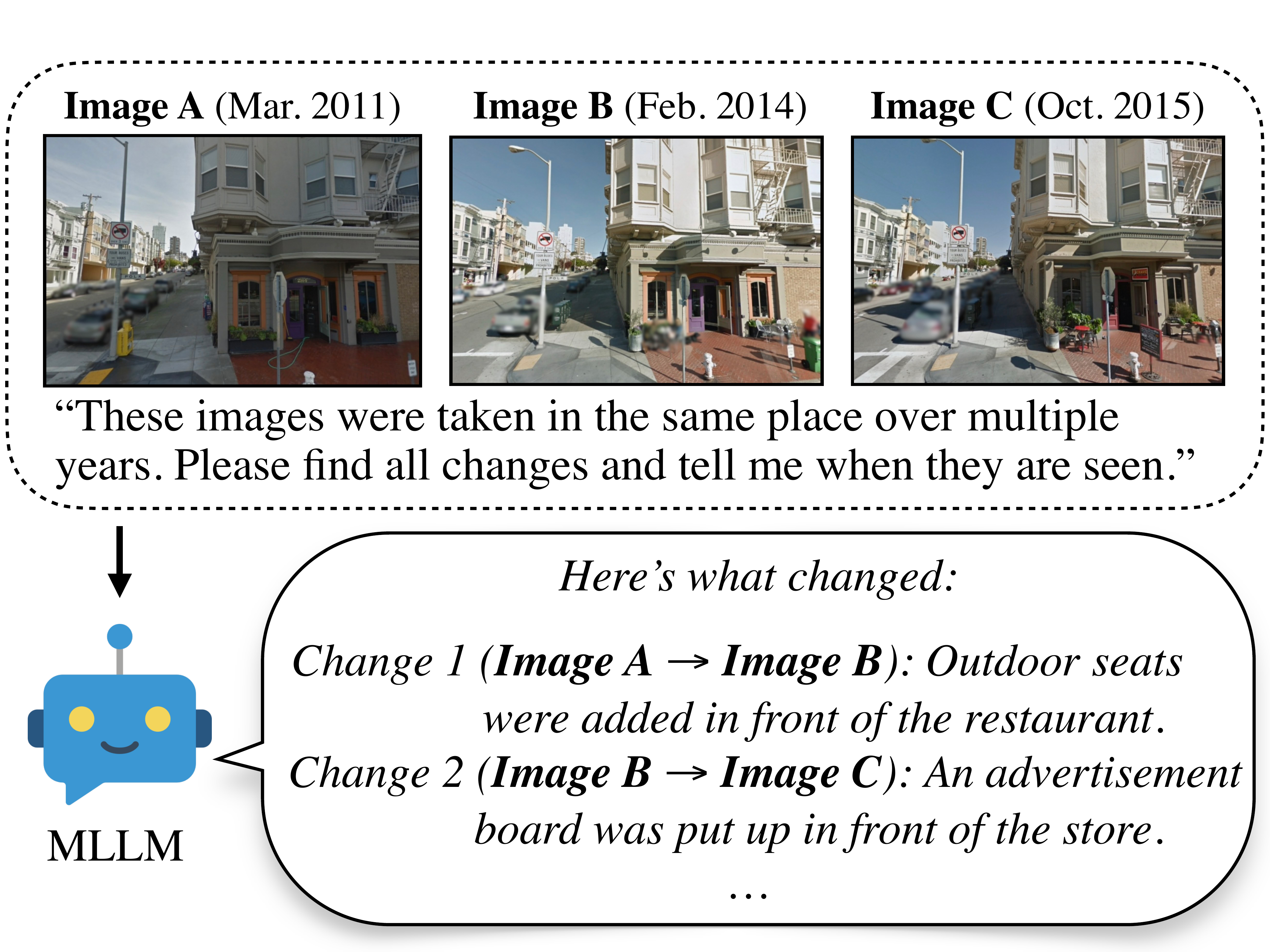}
\caption{
\textbf{Using MLLMs for Visual Change Detection}. 
Given a set of images captured from nearby views at different timestamps (top row), we use an MLLM as a visual analyst to detect changes.
}
\label{fig:local_change}
\end{figure}

\subsection{Vision-Language Models in Visual Analysis}
The evolution of VLMs has significantly advanced visual analysis. Early models, such as image captioning systems~\cite{vinyals2015show, karpathy2015deep}, were limited to generating descriptions for individual images, restricting analyses to small-scale tasks like captioning differences between pairs of images~\cite{park2019robust, kim2021agnostic,yao2022image, chang2023changes}.
Models like CLIP~\cite{radford2021learning}, ALIGN~\cite{jia2021scaling}, and their variants~\cite{zhu2022gsclip,zhang2024discover,csurka2024could} have also been used in various visual analysis tasks.  However, as VLM encoders without generative capabilities, they fall short for tasks requiring detailed image descriptions and reasoning.

The advent of large-scale VLMs, from Flamingo~\cite{alayrac2022flamingo}, PaLI~\cite{chen2022pali}, and BLIP-2~\cite{li2023blip}, to powerful new  models like Gemini~\cite{team2023gemini,team2024gemini} and ChatGPT~\cite{achiam2023gpt}, affords new opportunities for visual analysis.
These models integrate the expansive understanding of large language models with visual inputs, enabling sophisticated visual reasoning and generation.
While LLMs have been employed in large-scale text analyses~\cite{zhong2023goal, liang2024can, singh2024rethinking}, large-scale visual analyses using large VLMs remain scarce. A recent method, VisDiff~\cite{dunlap2024describing}, uses MLLMs to find differences between datasets of a few thousand images, but does not scale to our problem due to dataset size and differing objectives.
Also, their verification step is largely based on CLIP embeddings that prove to be unsatisfactory in our task.
One concurrent work, TeDeSC~\cite{liu2024organizing}, applies VisDiff's framework to semantic clustering tasks in relatively small datasets, organizing images based on text descriptions to find groups. 
Another recent work~\cite{zhong2024explaining} introduces a general analytical framework based on VLMs, utilizing creative language predicates as interpretable hypotheses. 
This approach, however, does not directly apply to our task due to the intractably large hypothesis space and the inability of LLMs to effectively predict them.
Additionally, completing their analysis requires labels for regression, equivalent to annotating ``is an image in a trend?'', which is extremely challenging in our context.
\section{Massive-Scale Image Analysis with MLLMs}
\label{sec:method}

\subsection{Task: Open-Ended Urban Imagery Analysis}
\label{sec:task}

Our work uses visual analysis to make open-ended discoveries from massive collections of images without labels.
We use two image collections (New York City and San Francisco) provided by Google Street View with permission, each with 20M posed and timestamped images captured 
at street level from 2011 to 2023.

Our objective is to identify and describe trends of visual changes over time in these image collections, with interpretable text descriptions accompanied by visually grounded evidence.
By \emph{visual change}, we refer to a semantic change event that occurred at a specific imaged location between two discrete times (corresponding to times when that location was imaged).
This query is open-ended because the event can be of any nature and happen to any kind of subject.
A \emph{visual trend} is a group of similar changes that have occurred at least $N$ times across different locations within a certain period, where $N$ is a hyper-parameter determining the minimum frequency for a trend to be considered significant.
Moreover, we desire the most detailed and specific insights we can find that still occur at least $N$ times, to gain a deeper understanding of subtle urban transformations.

\subsection{Using MLLMs for Massive Image Analysis}
\label{sec:image_analysis}

We aim to leverage the powerful analytical capabilities of MLLMs at large scales.
However, the na\"ive approach of feeding all the data to MLLMs does not work;
it is not possible to feed tens of millions of images into an MLLM due to context size limitations (e.g., Gemini-1.5 Pro~\cite{reid2024gemini}'s limit is 2 million tokens, which corresponds to at most $8$k images).
Instead, we decompose massive dataset queries into a set of local queries that can be answered independently by an MLLM. 
Then, we combine the answers to those local queries into an answer about the whole dataset.
This bottom-up approach is appropriate because there are natural ways to decompose an image dataset into small sets of images related to individual visual changes. During early experiments, we discovered that MLLMs are very effective (see Sec.~\ref{sec:change_det_eval}) at describing subtle changes within a temporal image sequence captured from nearly the same perspective (e.g., those shown in~\Fig{local_change}).  MLLMs not only detect small changes within such sequences, but also effectively ignore 
many common trivial changes (variations in time of day, season, etc.) and focus instead more on consistent semantic content that might be of interest to a person.
They also identify where in the sequence the visual changes occur in addition to producing detailed text descriptions of them (e.g., ``the corner store gets a new `for sale' sign on the window between the third and fourth image''). 

With this key insight regarding the capabilities of MLLMs, we design a two-step analysis system consisting of local change detection followed by trend discovery.
In both steps, we harness the analytical power of MLLMs, and design careful solutions to effectively process and analyze the data. 
The following sections describe each step in detail.

\subsection{Using MLLMs for Visual Change Detection}
\label{sec:local_change}
In the first step of our system, we detect local visual changes by analyzing small sets of images captured from approximately the same viewpoint at different times (\Fig{local_change} top row, see data processing details in Supplementary).
Since the images are time-stamped, we can sort them into chronological order at each location.
We feed every sorted image sequence into an MLLM and craft a query that, in essence, asks: 
1) what changes have occurred, and 
2) from which images these changes were detected. 
The output for the first question provides a text description for each detected visual change, while the second grounds these detections to specific visual evidence and timestamps.   
Please refer to the supplement for more details about image sequence preparation, MLLM prompting, and more example model outputs.

\begin{algorithm}[t]
\caption{Using MLLMs \protect\includegraphics[height=1.5em]{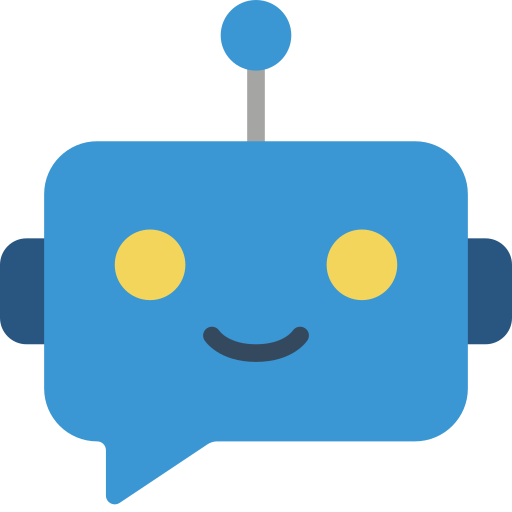}\ for Trend Discovery}
\label{alg:trend_discovery}
\begin{algorithmic}[1]
\Require Trend $\mathbf{T}$, set of changes $\mathcal{C}$, parameters $k$, $N$
\Ensure Is $\mathbf{T}$ a real trend or not?
\State Encode $\mathbf{T}$ and each $c_i \in \mathcal{C}$ using \texttt{TextEmb}
\State Compute distances $d_i$ between $\mathbf{T}$ and each $c_i$
\State Select $k$ nearest $c_i$ based on $d_i$ as set $\mathcal{C}_k$
\State Use MLLM to determine if each $c_j \in \mathcal{C}_k$ belongs to $\mathbf{T}$
\State Let $N_{\text{pos}}$ be the number of positive responses
\If{$N_{\text{pos}} \geq N$}
    \State \Return Trend $\mathbf{T}$ is \textbf{positive}
\Else
    \State \Return Trend $\mathbf{T}$ is \textbf{negative}
\EndIf
\end{algorithmic}
\end{algorithm}

\subsection{Using MLLMs for Visual Trend Discovery}
\label{sec:trend_discov}
In the second step, we aim to analyze the set of local visual changes produced by local change detection to discover frequent visual trends occurring throughout a city. Specifically, we look for trends where similar changes occur at least $N$ times at different locations in the same time period.

At first glance, one may consider a simple solution of presenting all local change descriptions to an MLLM and asking for trends.
However, this is impractical for massive city-scale datasets. 
For instance, from a collection of 20 million images for one city, the first step of the system produces over 3 million local changes.
Analyzing them all at once would exceed the context window size of current MLLMs.

To address this issue, we designed a two-step process that 
1) produces visual trend proposals,
and then
2) verifies which proposed trends are supported by at least $N$ visual changes. 
The first proposal step is relatively simple.
We embed the text description for every visual change into a vector space
and use Canopy Clustering~\cite{mccallum2000efficient} with a tolerant loose threshold to find potential cluster proposals.

The second verification step is more challenging.
An easy and efficient approach would be to assign each visual change to a trend proposal based on its proximity in an embedding space.
However, we find that this approach performs poorly because distances in embedding spaces are not distinctive enough to capture the subtle similarities and differences between visual changes required for our application. 
Image embeddings may encode both change-relevant and change-irrelevant content, and text embeddings may fail to capture subtle wording differences (e.g., adding ``not'') or different conceptual granularities (e.g., ``a storefront changed its business'' and ``a pizza shop changed into a café'' are difficult to cluster because they represent different levels of specificity).
Moreover, there is no universal embedding distance threshold to classify whether a change belongs to a trend, making it difficult to achieve consistent accuracy.

An alternative, albeit impractical, approach would be to feed every detected local change and every proposed trend to an MLLM and ask it whether the change fits within the trend (yes or no).
This approach is compelling because we find that MLLMs provide remarkably high accuracy for this small task. 
However, the downside of this approach is efficiency. 
Verifying 200 trend proposals exhaustively against 3 million changes using MLLMs would take more than one year.

To address this issue, we propose a hybrid algorithm which combines the efficiency of text embeddings with the precision of MLLMs, as shown in Alg.~\ref{alg:trend_discovery}.
Since we only need to find at least $N$ changes that belong to a trend to confirm it, we first use text embeddings to sort all changes based on their similarity to the trend proposal.
Next, we use the MLLM to verify only the top $k>N$ nearest neighbors. 
If $N$ positive matches are found among these $k$, the trend is confirmed. 
We set $k=1500$ for $N=500$ in our experiments, reducing the MLLM inference cost by $2000\times$ and making the process computationally feasible.
\section{Evaluations}
\label{sec:eval}
In this section, we evaluate Visual Chronicles against alternatives as well as design choices made in Visual Chronicles. We use Gemini-1.5 Pro~\cite{reid2024gemini} as the MLLM throughout the paper, for its impressive long-context multimodal analytical capability.
We collect a small set of data for evaluation.

\subsection{Comparisons to Trivial Use of MLLMs}
\label{sec:mllm_compare}
We first investigate how well our system works in comparison to na\"ive solutions using MLLMs trivially.

\medskip
\noindent \textbf{Can an MLLM predict similar visual trends without looking at any images?}
In the first experiment, we investigate how the visual trends produced by our proposed method compare to those produced directly by an MLLM when prompted only with text (i.e., without providing any images as context).
We acknowledge that this baseline is not apples-to-apples since our method can access lots of extra data and use significantly more compute.
However, it addresses an important question: ``Are MLLMs already knowledgeable enough to list visual trends on their own?''
To perform this comparison, we asked Gemini-1.5 Pro and GPT-4o to predict visual trends over the last decade using various prompt strategies.
We found that MLLMs' answers are usually abstract,
\eg ``Increased focus on sustainability'',
and general without intriguing details,
\eg ``store closures''.
Please see complete answers in Supplementary.
In contrast, our method identifies specific and subtle trends,
\eg addition of solar panels in \Fig{totd_sf},
more nuanced,
\eg closures of grocery stories in \Fig{retail_store},
and sometimes unexpected trends, 
\eg an overpass painted \emph{blue} in Fig.~\ref{fig:teaser},
all supported by visual evidence.
These visual trends provide a more detailed and verifiable understanding of changes in the city than the broad abstract ones provided by the MLLM on its own.

\medskip
\noindent \textbf{Can an MLLM process all the images in a massive collection in one query?}
In the second experiment, we investigate the limits of using MLLMs to predict visual trends with a single query using a set of images as context.
For this test, ideally, we'd provide the entire massive image collection of millions of images as context to the MLLM, and then ask ``what are the visual trends?''
However, this is not possible, since MLLMs have limits on the number of tokens that can be provided as context ($8$K images for the MLLM with largest limit~\cite{reid2024gemini}).
So, instead we performed tests with randomly sampled subsets of $8$K images out of $20$M in NYC as context, and take the union of answers from all subsets as the final result.
The results however are not satisfactory--due to the severe under-sampling of images in each batch,
only the extremely frequent trends can be found, \eg the addition and removal of scaffoldings in NYC which our system also identifies.
Additionally, this solution only finds less than $10$ trends throughout a whole city even after union of multiple sub-samples.
See complete answers in Supplementary.

\subsection{Local Change Detection Evaluation}
\label{sec:change_det_eval}
Next, we quantitatively evaluate the design choices made in Visual Chronicles, starting from local change detection.
To understand how good Visual Chronicles is at local change detection, we collect a dataset of $200$ locations with $3036$ images in total.
For each location, we manually label the changes, including between which $2$ consecutive images each change happens and the description of changes.

We compare our method with prior unsupervised image analysis tools, including image features such as HoG features~\cite{doersch2015makes} and Color Histograms (C-Hist)~\cite{torralba2011unbiased}, and semantic features such as CLIP~\cite{radford2021learning} and NV-Emb~\cite{lee2024nv}.
These methods first featurize every image in the image sequence for each location and measures the feature distance between consecutive images as an indicator for potential change.
For NV-Emb, we use Gemini~\cite{reid2024gemini} to caption each image and embed the caption text.
Additionally, we compare ours with an unsupervised remote sensing method (R-Sensing)~\cite{melgani2002unsupervised}.
Note that other learning-based approaches~\cite{gebru2017using} requires labelled training data, hence are inapplicable.

To have a fair comparison between MLLMs and baselines that only predict the change likelihood between two images, we use a classification metric, Average Precision (AP).  
For any consecutive image pairs in our test set, our human labels give us binary ground truth on whether there is a change. We then compute the AP on this binary classification task for all methods. Note that if there are multiple changes between a pair of images, we only count the pair once.
Additionally, to further measure MLLMs' performance, we also evaluate
the precision of its prediction by manually labelling if each detected change is true,
that is if the text is actually describing a change
and the change align with the corresponding evidence images,
as well as its recall by manually labelling if each human labelled change is covered in MLLMs' answers.

The results in \Tab{change_det} show that the MLLM (Gemini) is significantly better than the two methods based on embedding distances -- i.e., by more than 50 AP percentage points.  Overall, we find that the MLLM is remarkably good at this task -- its precision is $81.34\%$ and its recall is $89.87\%$.

\begin{table}
\centering{
\resizebox{\columnwidth}{!}{
\begin{tabular}{lcccccc}
\toprule
 & HoG~\cite{doersch2015makes} & C-Hist~\cite{torralba2011unbiased} & R-Sensing~\cite{melgani2002unsupervised} & CLIP~\cite{radford2021learning} & NV-Emb~\cite{lee2024nv} & Gemini~\cite{reid2024gemini} \\
\midrule
\textbf{AP} & $16.44 \%$ & $16.76 \%$ & $18.51 \%$ & $26.52 \%$ & $23.75 \%$ & $\mathbf{76.56 \%}$ \\
\bottomrule
\end{tabular}
} 
} 
\caption{\textbf{Evaluating Visual Change Detection}.
Gemini detects visual changes with significantly higher average precision (AP) than baselines based on image features (HoG, C-Hist), remote sensing (R-Sensing), and semantic embeddings (CLIP, NV-Emb).
}
\label{tbl:change_det}
\end{table}

\subsection{Trend Discovery Evaluation}
\label{sec:trend_discov_eval}
\begin{table}
\centering{
\resizebox{0.96\columnwidth}{!}{
\begin{tabular}{lccc}
\toprule
\textbf{Representation} & \textbf{Type} & \textbf{Scalable} & \textbf{Average Precision} \\
\midrule
    Random & --- & --- & $47.70 \%$ \\
\midrule
    CLIP & ImgEmb & \greencmark & $54.78 \%$ \\
    NV-Emb & TxtEmb & \greencmark & $73.13 \%$ \\
\midrule
    Gemini & MLLM & \redxmark & $\mathbf{86.63 \%}$  \\
\bottomrule
\end{tabular}
} 
} 
\caption{\textbf{Evaluating Trend Discovery}.
The MLLM discovers trends with significantly higher average precision (AP) than baselines based on embedding distances, but is not scalable due to expensive inference.
}
\label{tbl:trend_discov}
\end{table}
\begin{table}
\centering{
\resizebox{.98\columnwidth}{!}{
\begin{tabular}{lcccc}
\toprule
\textbf{Method} & AllTrue & NV-Emb & RandMLLM & Hybrid (\textbf{Ours}) \\
\midrule
\textbf{Acc.@$50$} & $72.7\%$ & $77.9\%$ & $31.8\%$ & $\mathbf{93.9\%}$ \\
\textbf{Acc.@$100$} & $54.1\%$ & $69.6\%$ & $49.9\%$ & $\mathbf{94.6\%}$ \\
\textbf{Acc.@$200$} & $28.9\%$ & $81.8\%$ & $74.9\%$ & $\mathbf{98.3\%}$ \\
\bottomrule
\end{tabular}
} 
}
\caption{\textbf{Evaluating Change-Trend Classfiers}.
AllTrue is predicting positive for all.
Our hybrid algorithm is significantly more accurate at verifying change-trend pairs than scalable alternatives.
}
\label{tbl:hybrid}
\end{table}

\paragraph{Trend discovery is not trivial for MLLMs.}
We first evaluate a trivial solution of feeding all change descriptions to MLLMs and asking for trends, to the extent this is possible. Current context limits can support at most $50$K changes at once, out of $3$M changes per city (less than 2\%).
We therefore subsample sets of $50$K changes and take the union over multiple sets, but the answers are not satisfactory.
Similar to feeding subsampled image collections to MLLMs, only very few trends can be identified and they are dominated by extremely frequent trends, \eg scaffolding and construction.
See the supplementary for more details.

\paragraph{MLLMs are good at verifying trends, but are not scalable.}
Next, we test MLLM trend verification capabilities.
We sample a collection of $50$ cluster proposals each with $40$ change candidates to verify.
This makes $2000$ trend-change pairs for the binary classification task ``does this change belong to this trend?''.
We use human labellers to manually label these pairs.
To ensure more balanced label distribution, we over-label more randomly sampled pairs and subsample aforementioned $2000$ pairs to balance positives and negatives.
We compare MLLMs with SOTA image embedding (CLIP) and text embedding (NV-Emb), as well as a trivial solution that produces uniformly random binary predictions.
As in the previous test, we use average precision as the metric. 
Our results in \Tab{trend_discov} show that the MLLM (Gemini) significantly outperforms the methods based on embeddings, by a margin more than $13\%$.
However, MLLMs are expensive in inference time.
Even with $64$ MLLMs running in parallel, it would take $380$ days to verify $200$ trend proposals against $3$M changes (0.6B queries at 3.5s/query/MLLM), while embedding approaches can finish in less than a day with $64$ A100 GPUs.

\paragraph{Our hybrid solution is both accurate and scalable.}
At last, we verify performance of our hybrid solution.
We know that our solution reduces the required MLLM inference cost by $2000\times$, making it sufficiently fast to verify trends, \eg $4.6$ hours to verify $200$ trends with the same $64$ parallel MLLMs.
So, the remaining question is whether it can discover trends accurately.
To test this, we sample $1000$ trend proposals, each with $5000$ changes to verify against.
We use an MLLM to exhaustively verify every trend-change pair, $5$M in total, and label each trend proposal as true or false, based on whether the MLLM confirmed at least $N$ changes for it (shown as @$N$ in \Tab{hybrid}).  We then test our hybrid method against with this labeled set of trends to measure how accurately it classifies them as true or false, and report the binary classification accuracy.
We compare our hybrid solution with:
(1) A trivial approach that always predicts true,
(2) A pure text embedding approach (NV-Emb) where we use a threshold to binarize trend-change similarity and use the binarised label to decide if a trend proposal is true. We report the best accuracy among $1000$ grid-searched thresholds, and 
(3) Randomly selecting changes to verify (RandMLLM). Unlike our text embedding sorted verification, this solution \emph{randomly} picks the same amount of changes as our solution to verify for each trend proposal.
In \Tab{hybrid}, we find that, compared to other scalable alternatives, our hybrid solution is significantly more accurate. It does not degrade much the accuracy from exhaustive MLLM inference while being much more efficient.
For the hyper-parameter $k$ in our solution, we use $k=3N$ and ablate the choice of $k$ in Supplementary.
\begin{figure*}[t]
\centering
\includegraphics[width=0.98\textwidth]{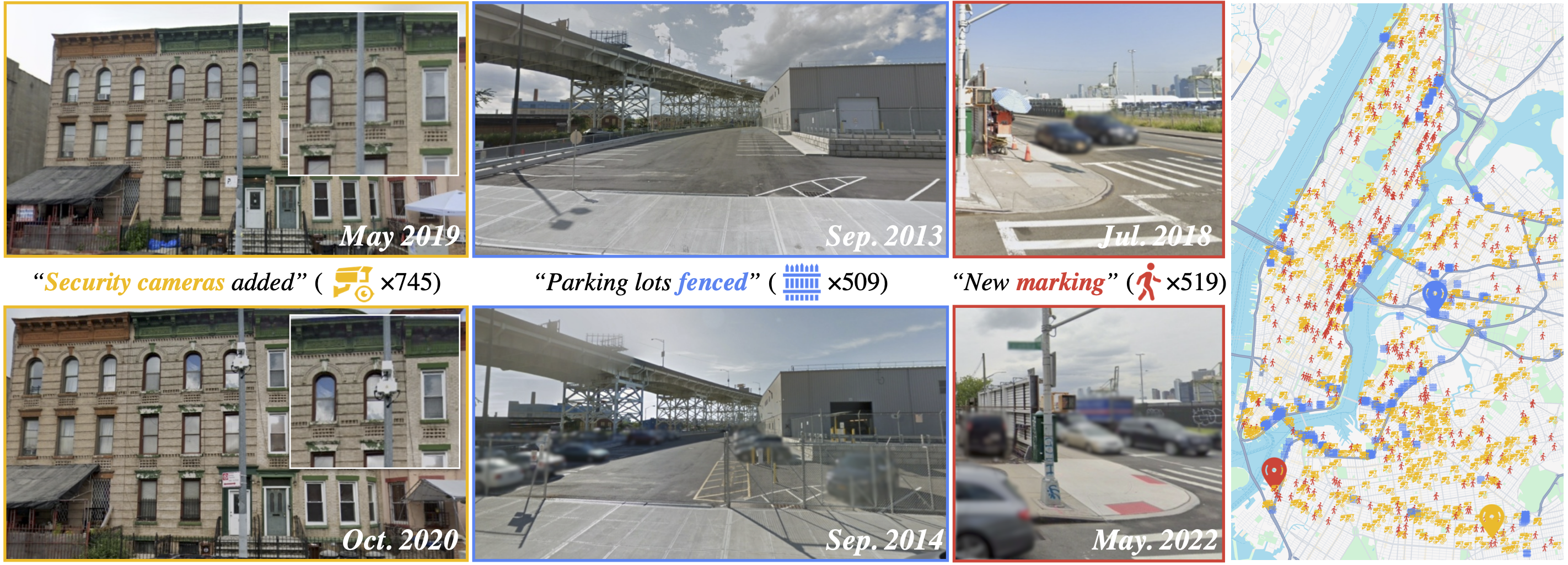}
\caption{
\textbf{Trends of the Decade in New York City}.
We discover interesting trends of visual change in NYC, including 
745 new security cameras, 509 new fences around parking lots, and 519 red ADA warning pads on sidewalks.  
For each example, the top row shows images before a change, bottom row shows after, and markers appear on the map where the changes occurred.
}
\label{fig:totd_nyc}
\end{figure*} 
\section{Applications}
\label{sec:application}

In this section, we apply our system
to study visual trends spanning a decade in two major US cities, San Francisco (SF) and New York City (NYC).
For each city, we collect 20M Street View images, captured from 2011 to 2023.
For NYC, the images are sampled from 883K different locations, with an average of 23 images per location.
For SF, we sample from 943K locations, with an average of 21 images per location.
Local change detection yields 2.9M change entries in NYC and 3.6M in SF.
For each city, we verify the most detailed 500 trend proposals.

\begin{figure*}[t]
\centering
\includegraphics[width=0.98\textwidth]{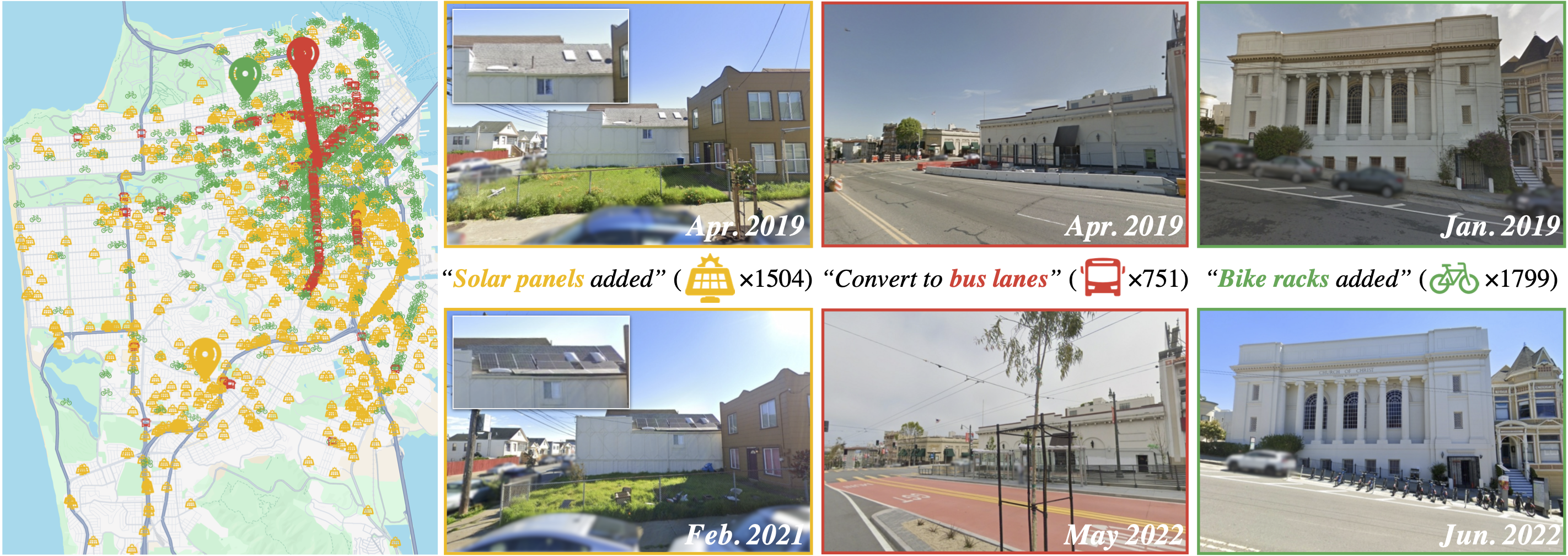}
\caption{
\textbf{Trends of the Decade in San Francisco}.
In San Francisco, we discover 1504 new solar panels (mostly in residential areas), 751 bus lanes conversions (mostly along a few major roads), and 1799 new bike racks (mostly near downtown).
}
\label{fig:totd_sf}
\end{figure*} 
\subsection{What are the trends of the decade?}
\label{sec:uncondition}
We first show the results of an open-ended attempt to discover ``what changes were frequently observed over a decade?'' in each city.  Unlike most prior work, these results are produced without any preconditioning.
That is, no external input was provided to guide the system to find specific types of changes, to avoid finding commonly occurring trivial changes (time of day, seasons, etc.), and to avoid semantically irrelevant changes (e.g., presence of distractors).  
The notion of what types of trends are worth reporting is determined by the MLLM during our local change detection and trend discovery processes.

\Fig{totd_nyc} and \Fig{totd_sf} show a sampling of the visual trends found in NYC and SF, respectively (more appear in the Supplementary).
For each trend,
we can provide hundreds of image pairs as supporting visual evidence.  

Yet, we still must be careful when drawing conclusions. For example, in \Fig{totd_sf}, it appears many new solar panels were added near the raised freeway. In reality, it may be that roof-mounted solar panels are more likely to be visible in an image from a raised freeway than from ground level.  We must be careful to account for spatial and temporal image sampling biases when reviewing the results.
Despite this, we believe our system can be useful for discovering what changes happened, and where they happened, in an evidence-driven way that previously would not have been possible without an enormous amount of human effort.
\subsection{What happened in SF since 2020?}
\label{sec:time_condition}
Our system also supports searching for visual trends regarding a specific sub-period of interest.
To demonstrate, we showcase a time-conditioned search for the range 2020-01-01 to 2022-12-31.
Here, we filter to changes that happened strictly within this period, i.e., both images (before and after) are captured within this period, and we rerun only the trend discovery step.
To encourage the system to find trends that might be unique to post-2020, we use a simple heuristic to sort trend proposals by their potential frequency differences between post-2020 years (2020--2022) and pre-2020 years (2017--2019).
We conduct this study on data from SF.

\paragraph{Outdoor dining.}
One notable trend is 
``The storefront added tables and chairs outside'', indicating the addition of outdoor dining, as shown in \Fig{teaser}.
We observe this occurrence 1,482 times in 2020-2022 and 668 times in 2017--2019.

\paragraph{An Overpass Painted Blue.}
To our surprise, \Fig{teaser} shows the visual trend ``The support of the overpass was painted blue.'', observed $481$ times since 2020 ($5$ times before 2020). An internet search reveals the likely explanation-- San Francisco settled on `Coronado Blue' for its \$31 million paint job of the Central Freeway (\href{https://sfstandard.com/2022/10/25/central-freeway-31-million-paint-job/}{SF Standard reporting}).
\subsection{Trends related to retail stores in NYC}
\label{sec:semantic_condition}
Visual Chronicles also supports searches for trends relevant to a specific subject.
We provide an example study of retail stores in NYC.
Similar to \Sec{time_condition}, we can re-use the local changes detected in \Sec{uncondition}, and this time filter them by semantic relevance.
To compute semantic relevance to retail stores, we first embed the sentence, ``A retail store has changed''. Then, similar to our hybrid trend verification solution, we pick the nearest $500$K changes in NYC and use an MLLM to check that they are indeed changes relevant to retail stores.
After filtering, we run the trend discovery stage, except we verify $150$ trends uniformly across trends of varying word count (rather than the lengthiest trends), hoping to find a variety of trends about retail stores.

\begin{figure}
\centering
\includegraphics[width=0.92\linewidth]{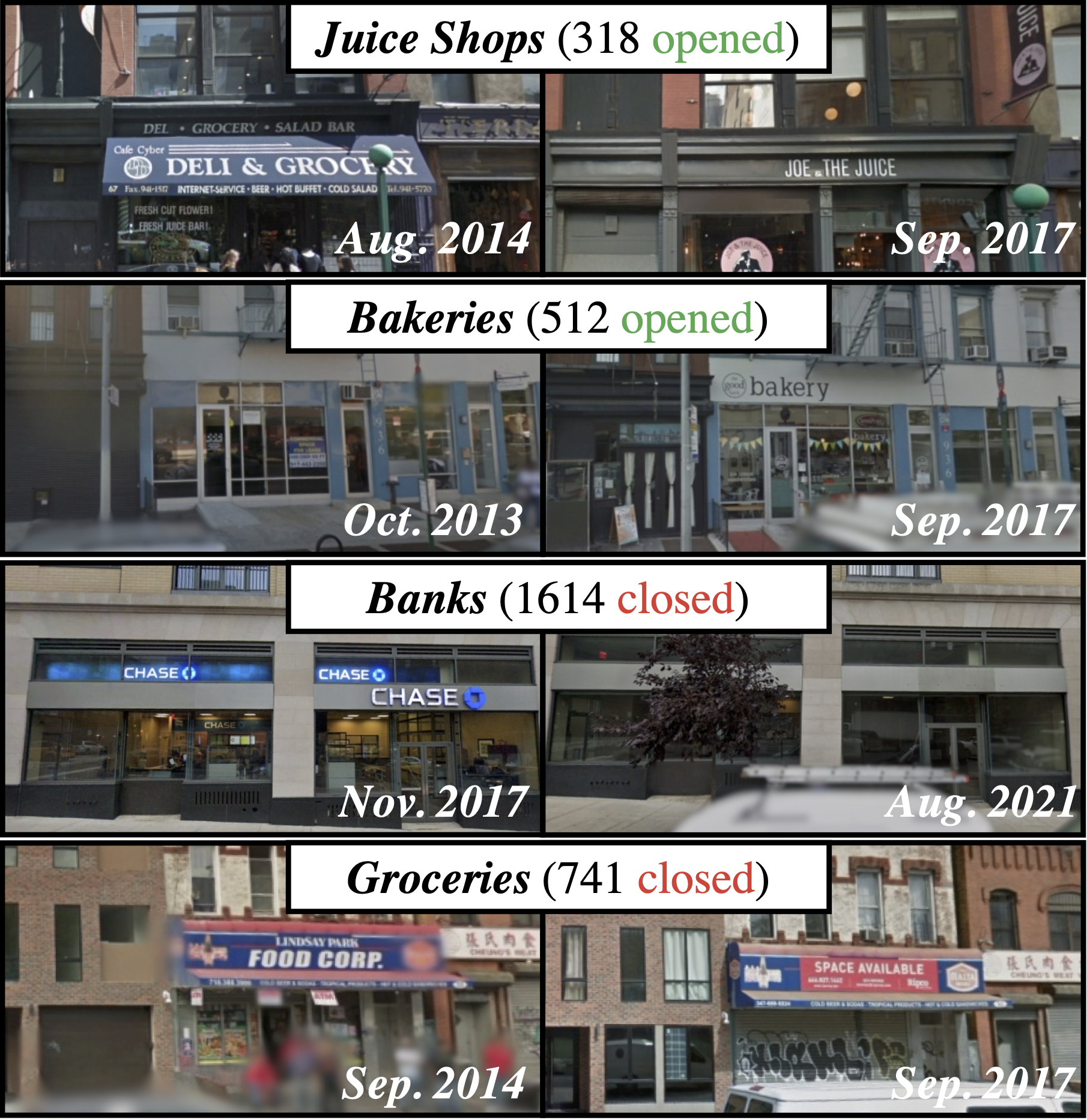}
\caption{
\textbf{NYC Retail Visual Trends}.
By conditioning on
``a retail store has changed,`` we are able to discover visual trends related to the opening and closing of types of retail stores in New York.
}
\label{fig:retail_store}
\end{figure}

\paragraph{Bakeries and Juice Shops.} Seen in \Fig{retail_store}, one type of trend that arises within this search is the opening of various shops, including ``There is a bakery newly opened at the storefront'' and ``A juice shop opened at the storefront''.

\paragraph{Groceries and Bank Branches.} Opposite to the opening of certain stores, the closing of certain stores also happened in 2011--2023 (\Fig{retail_store}).
Among our trends, we find that many grocery stores and bank branches closed in NYC.

\subsection{A Non-Temporal Query: ``Unusual Things''}
\label{sec:alternative}
\begin{figure}[t]
\centering
\includegraphics[width=\linewidth]{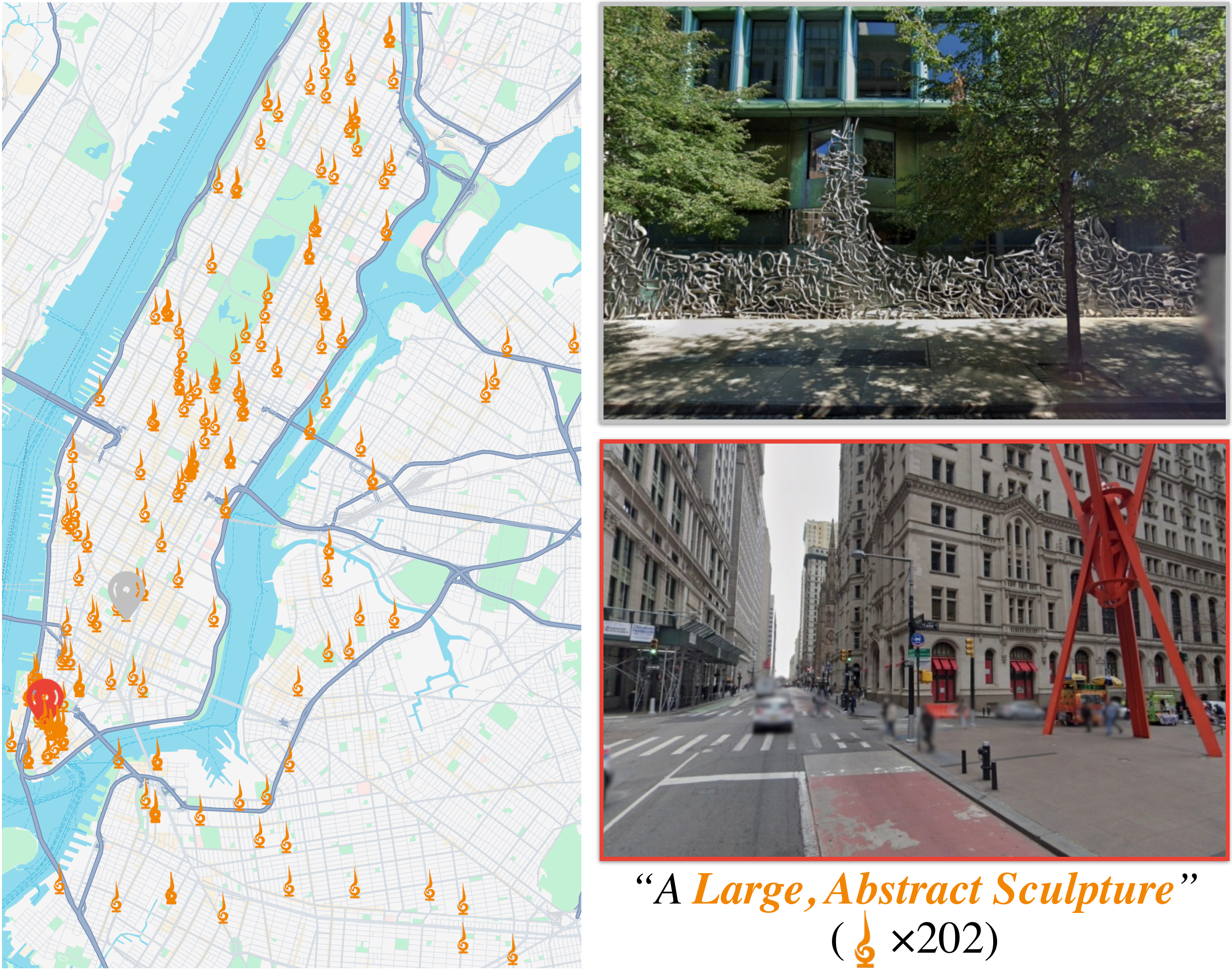}
\caption{
\textbf{Unusual Things on the Streets of NYC}.
Beyond temporal queries, our system can also be applied to analyzing individual images, for non-temporal open-ended queries.
Here we show our findings from the individual image query ``What are the unusual things in the image?''.
We find there are many \emph{unusual} ``large, abstract'' sculptures in NYC.
}
\label{fig:odd_things}
\end{figure}
Our system also supports non-temporal open-ended queries.
As one example, we investigated searching for ``unusual things'' that actually happen a lot in a city.
To do so, we simply changed the first step of our system to send just a single image to the MLLM and ask it ``what are the unusual things in the image?.''
The MLLM then produces text describing the local unusual things,
which will be processed by the second trend discovery step without change.
Applying this process to a large set of images in NYC yields a cluster of images showing ``a large, abstract sculpture'' (\Fig{odd_things}), among other interesting clusters.
\section{Conclusions, Limitations, and Future Work}
\label{sec:conclusion}

This paper explores how to use MLLMs to analyze
a massive image collection without labels to find trends with open-ended semantic queries.
In addition to introducing this novel task, we carefully design a novel system effectively leveraging the strong capabilities of MLLMs for large scale analyses.
Experiments show the proposed system is able to
discover fascinating trends automatically.

This paper is the first to analyze image datasets with
millions of images using
MLLMs, and so it should be considered an initial proof-of-concept,
rather than an exhaustive analysis of all possible design choices.
Accordingly, there are many limitations. 
One key avenue for future work is to integrate MLLM analysis 
over large-scale, distributed imagery
into a robust statistical framework that takes into account biases, e.g., due to uneven sampling of imagery and inaccuracies or bias in MLLM results --- our work is a scalable way to generate hypotheses from vast amounts of visual data, but the methods we devise could also be developed into tools that allow for more rigorous conclusions, grounded in evidence.
Another fruitful direction would be to apply our methods to other kinds of large visual datasets, including video (e.g., archival news footage), and to other kinds of queries,
like
``find interesting styles.''
In any
case, we are likely near the beginning of a long line of research
related to using MLLMs for analyzing massive image collections, and
this paper takes a small step in that direction.

\paragraph{Acknowledgements}.
Thanks to Jiahui Lei, Anh Thai, Jiapeng Tang, Linyi Jin, Luming Tang, Rundi Wu, Ian Huang, Colton Stearns, Francis Engelman, Manu Gopakumar, Suyeon Choi, Haley So, Richard Tucker, Abhijit Kundu, Jonathan Barron, Glenn Entis, and David Salesin, for their comments and constructive discussions; to Abhijit Kundu, William Freeman, and John Quintero for helping review our draft; G.W. was in part supported by Google, Samsung, and Stanford HAI. B.D. was in part supported by a Qualcomm Innovation Fellowship.
%
{
    \small
    \bibliographystyle{ieeenat_fullname}
    \bibliography{main}
}

\clearpage
\setcounter{page}{1}
\maketitlesupplementary

In this supplementary, we provide 
more application results including an example of a non-temporal query and a spatial insight (\Sec{more_app}),
more details about our data, system, and evaluations (\Sec{more_details}),
more evaluations including ablations and complete results of MLLM baselines on the hyper-parameters of Visual Chronicles (\Sec{more_eval}),
and potential societal impacts of Visual Chronicles (\Sec{social_impact}).

\section{Extended Application Results}
\label{sec:more_app}

\subsection{Where are new buildings in NYC?}
\label{sec:new_buildings}
\begin{figure}[t]
\centering
\includegraphics[width=0.98\linewidth]{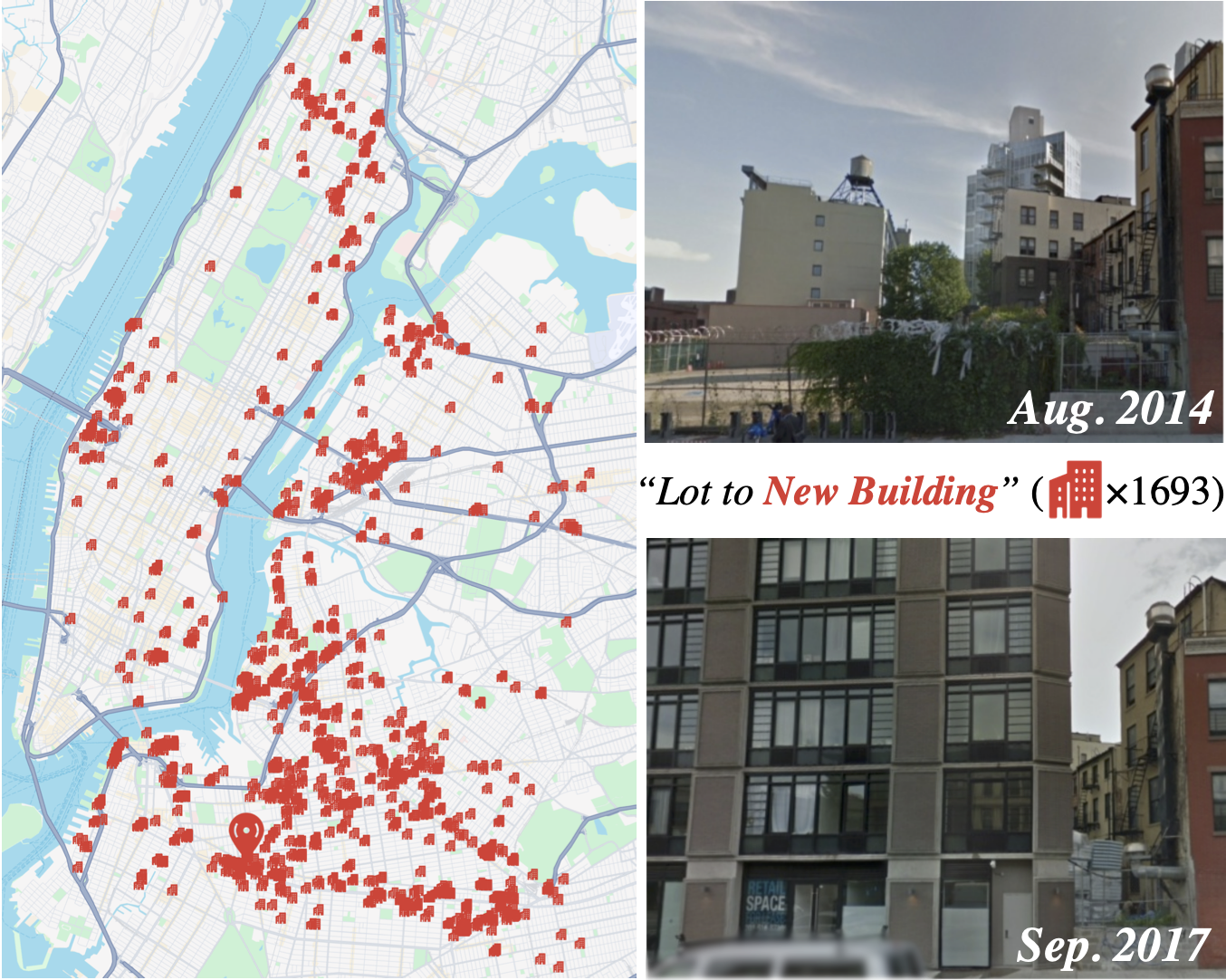}
\caption{
\textbf{Where are new buildings in NYC?}
We plot all locations where we observe the trend, ``A lot now has a multi-story building constructed on it'', in NYC.
We find notable clustering in Brooklyn, Long Island City, and Hudson Yards.
}
\label{fig:new_buildings_nyc}
\end{figure}
Aside from the analysis for specific temporal period of interest (\Sec{time_condition}) or semantic subject of interest (\Sec{semantic_condition}),
Visual Chronicles can also reveal insights about the spatial domain, \eg by visualising the spatial distribution of a trend.
In particular, in this study, we showcase the spatial distribution of one of the trends of the decade in NYC (\Sec{uncondition}),
namely ``A lot now has a multi-story building constructed on it''.
As shown in \Fig{new_buildings_nyc}, we find that more new buildings are observed in Brooklyn and Long Island City, compared to sparser, scattered points in Manhattan.
Additionally, the well-known Hudson Yards development~\cite{enwiki:1247695752} appears as a rare spatial cluster in Manhattan.

\subsection{Many Graffiti Observed Since 2020}
\label{sec:graffiti}
\begin{figure}[t]
\centering
\includegraphics[width=0.98\linewidth]{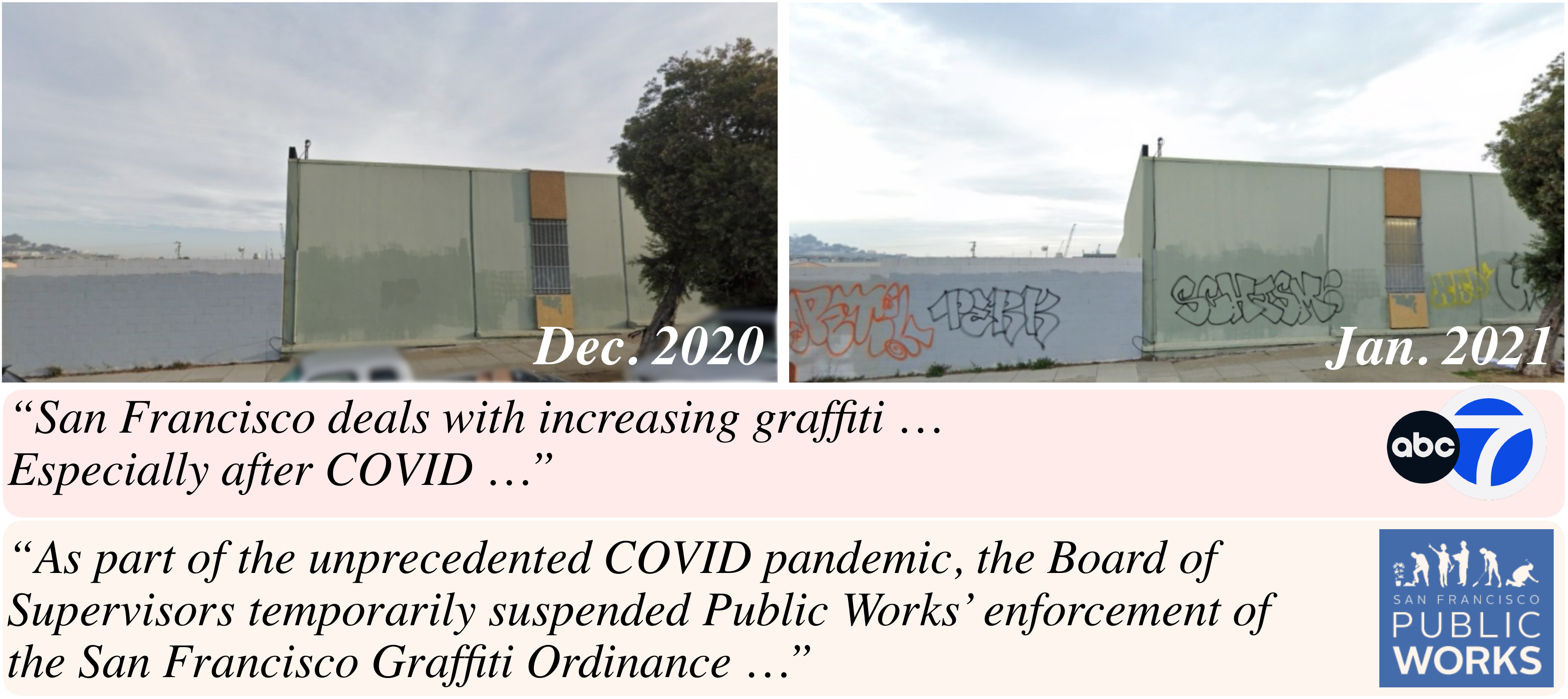}
\caption{
\textbf{Many Graffiti Observed, 2020--2022}. In SF since 2020, many new graffiti are observed.
A quick search on the Internet reveals relevant reports in local news and a local policy change.
The enforcement has since resumed.
}
\label{fig:graffiti}
\end{figure}
As shown in \Fig{graffiti}, our system also identified the trend ``New graffiti were added to the wall'' as a significant trend since 2020.
Such changes were spotted $3152$ times post-2020 and $1150$ times pre-2020. As reported by \href{https://abc7news.com/graffiti-in-san-francisco-tagging-vandalism-street/13801629/}{ABC News}, during COVID San Francisco paused enforcement of local regulations requiring building owners to remove graffiti promptly.
Yet, this is an example of why we must be careful when drawing conclusions from this data, as at least two hypotheses seem compatible: 1) graffiti tagging became more popular, or 2) graffiti was removed less frequently, increasing its likelihood of being observed in our data.

\subsection{Additional Trend Results}
\label{sec:more_trends}
\begin{figure*}[t]
\centering
\includegraphics[width=0.98\textwidth]{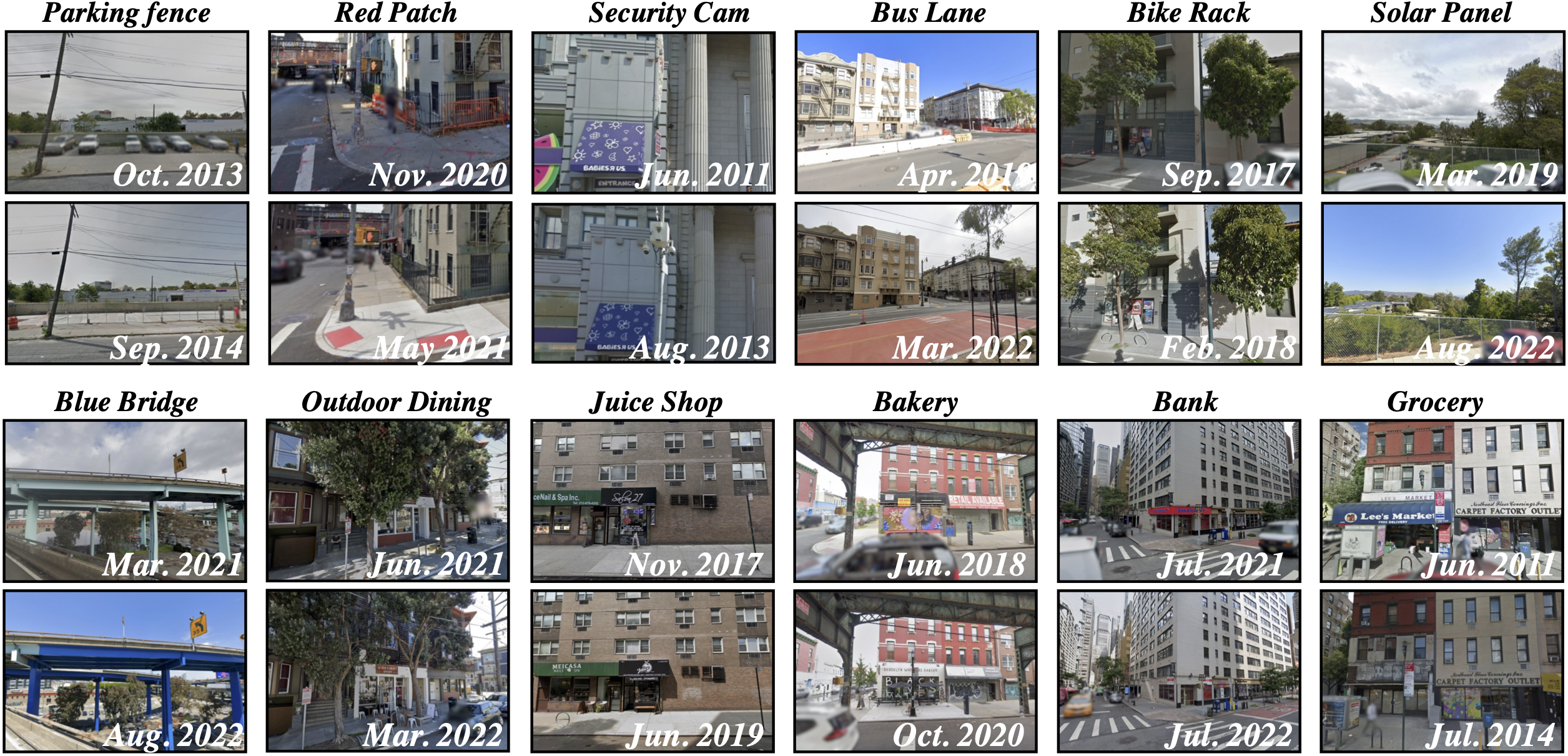}
\caption{
\textbf{More Images of Shown Trends}.
}
\label{fig:more_imgs}
\end{figure*}

\paragraph{More Images of Shown Trends.}
In previous sections, we have shown example trends discovered in Visual Chronicles, each evidenced by a pair of images.
In \Fig{more_imgs}, we further provide an additional pair of images illustrated visual evidence for each of these previously shown trends, demonstrating the variation in the visual data.

\begin{figure*}[t]
\centering
\includegraphics[width=0.98\textwidth]{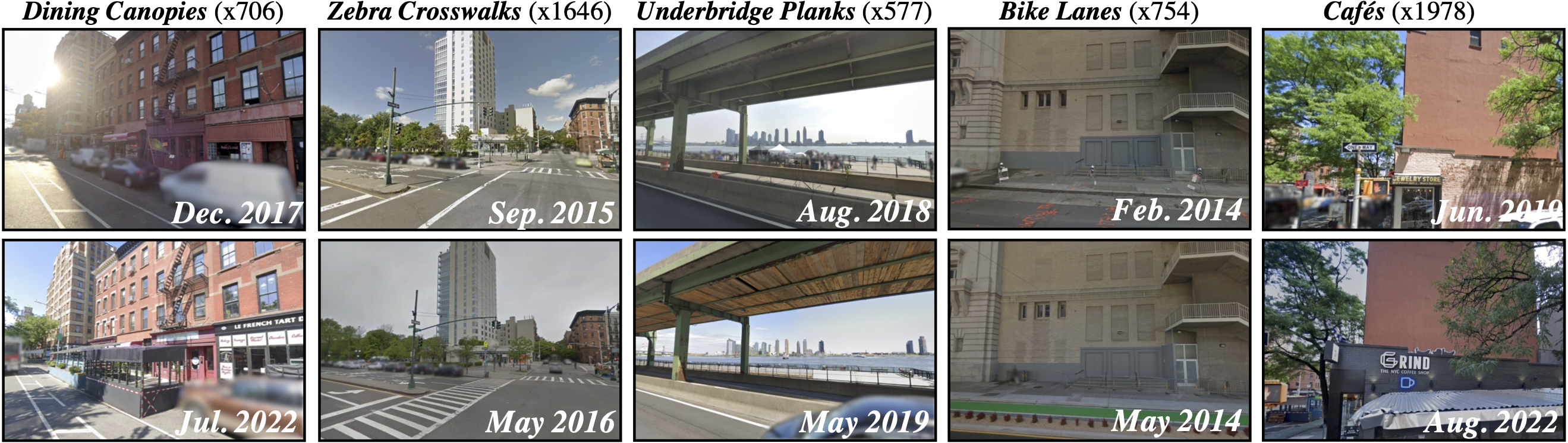}
\caption{
\textbf{More Trends in Images}.
}
\label{fig:more_trends_img}
\end{figure*}

\paragraph{More Trends in Images.} In addition, we show in \Fig{more_trends_img} more trends found using Visual Chronicles along with accompanying visual evidence.
Such trends include a specific ``canopied outdoor dining'' trend in NYC, a ``new zebra crosswalk'' painted on the street, an added ``under-bridge wooden plank'', an added green ``bike lane'', and the opening of ``new cafés''.

\begin{figure*}[t]
\centering
\includegraphics[width=\textwidth]{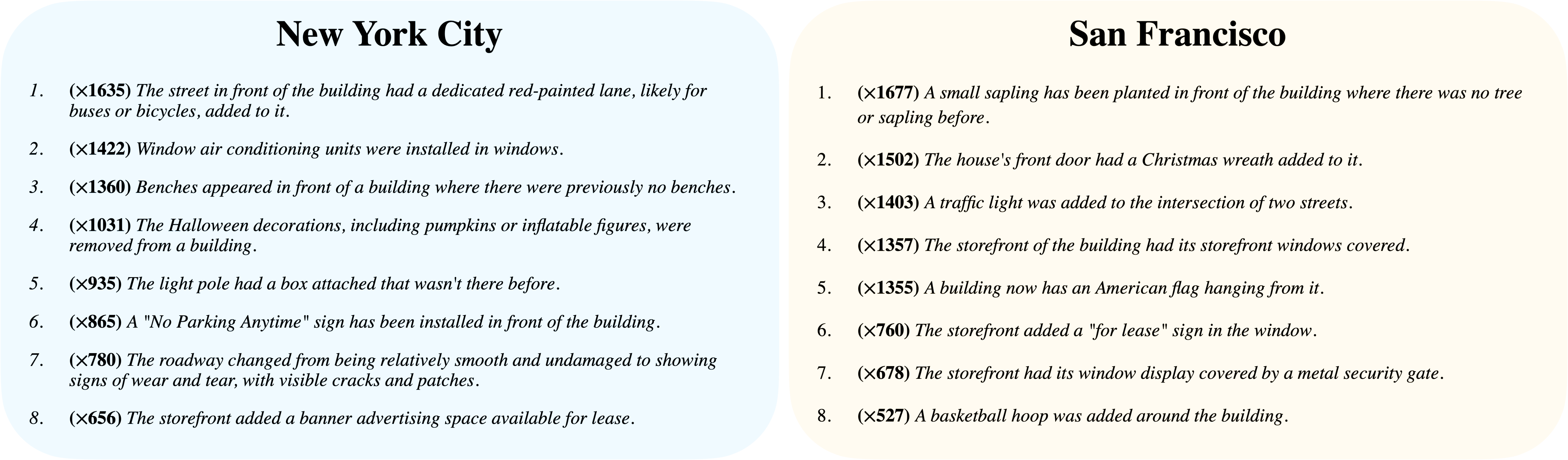}
\caption{
\textbf{More Trends in Text}.
}
\label{fig:more_trends_text}
\end{figure*} 
\paragraph{More Trends in Text.} 
In previous figures, we have shown all trends with visual evidence.
For visualization purpose, the shown text is shortened descriptions of the exact trend found in Visual Chronicles.
In \Fig{more_trends_text} and~\Tab{trends_for_shown_figures}, we first show a list of complete text for all the trends we have shown.
Moreover, we show an addition set of found trends also in complete text descriptions.
\begin{table*}[t]
\centering
\begin{tabular}{c|p{0.8\textwidth}}
\hline
& \textbf{Trend Description} \\
\hline
\multirow{2}{*}{\textbf{\Fig{teaser}}} 
    & (\textbf{$\times$1483}) \textit{The storefront added tables and chairs outside.} \\
    & (\textbf{$\times$481}) \textit{The support of the overpass was painted blue.} \\
\hline
\multirow{3}{*}{\textbf{\Fig{totd_nyc}}} 
    & (\textbf{$\times$745}) \textit{Security cameras became visible on the light poles.} \\
    & (\textbf{$\times$509}) \textit{The parking lot in front of the building now has a fence enclosing it.} \\
    & (\textbf{$\times$519}) \textit{The crosswalk had its marking changed from white to red.} \\
\hline
\multirow{3}{*}{\textbf{\Fig{totd_sf}}} 
    & (\textbf{$\times$1504}) \textit{Solar panels were added to the roof of one of the buildings in the background.} \\
    & (\textbf{$\times$751}) \textit{The street has been reconfigured from a standard two-way configuration to a configuration with a dedicated bus lane.} \\
    & (\textbf{$\times$1799}) \textit{Bike racks were added in front of a building.} \\
\hline
\multirow{4}{*}{\textbf{\Fig{retail_store}}} 
    & (\textbf{$\times$318}) \textit{A juice shop opened at the storefront.} \\
    & (\textbf{$\times$512}) \textit{There is a bakery newly opened at the storefront.} \\
    & (\textbf{$\times$1614}) \textit{The bank branch at the storefront closed.} \\
    & (\textbf{$\times$741}) \textit{The grocery store on the street closed.} \\
\hline
\multirow{1}{*}{\textbf{\Fig{new_buildings_nyc}}} 
    & (\textbf{$\times$1693}) \textit{A lot now has a multi-story building constructed on it.} \\
\hline
\multirow{1}{*}{\textbf{\Fig{graffiti}}} 
    & (\textbf{$\times$3152}) \textit{New graffiti were added to the wall.} \\
\hline
\multirow{5}{*}{\textbf{\Fig{more_trends_img}}} 
    & (\textbf{$\times$706}) \textit{An outdoor seating area with a canopy was added in front of the buildings.} \\
    & (\textbf{$\times$1646}) \textit{The crosswalk changed from having solid white painted lines to zebra-striped painted lines.} \\
    & (\textbf{$\times$577}) \textit{Portions of the underside of the overpass structure are covered with wooden planks, where it was primarily metal beams and supports before.} \\
    & (\textbf{$\times$754}) \textit{A green bike lane was added to the street in front of a building.} \\
    & (\textbf{$\times$1978}) \textit{The storefront changed into a cafe.} \\
\hline
\end{tabular}
\caption{\textbf{Full Texts for Trends in Various Figures}.}
\label{tbl:trends_for_shown_figures}
\end{table*}

\section{Extended Details}
\label{sec:more_details}

\subsection{Computation Costs.}
\label{sec:cost}
Our change detection step takes $<24$ hours for analysing $20$M images from $1$ city.
As in L$457$, our aggregation step takes $4.6$ hours for $200$ trends. 
Both steps use $64$ MLLMs running in parallel.
The total cost of MLLM inference is $<10K$\$, well within the budget of commercial city-scale analysis projects.

\subsection{Data Preprocessing}
\label{sec:dataset_preproc}
To produce the sets of local image sequences for the local change detection step of our system,
we first randomly sample a set of seed locations from the dataset.
For each seed location, we then sample a view direction that's likely to have buildings in view, based on the map.  Finally, for each panorama centered within $1.8$ meters of the seed location, we reproject the panorama to the sampled view direction as a pinhole image and add the resulting image to the set. 
For a city-scale dataset with 20M images (e.g., NYC), this process results in a manageable set of 10 to 40 images in each set at $\sim$900K sampled locations.

\subsection{Dataset Coverage}
\label{sec:dataset_details}
\begin{figure}[t]
\centering
\includegraphics[width=0.96\linewidth]{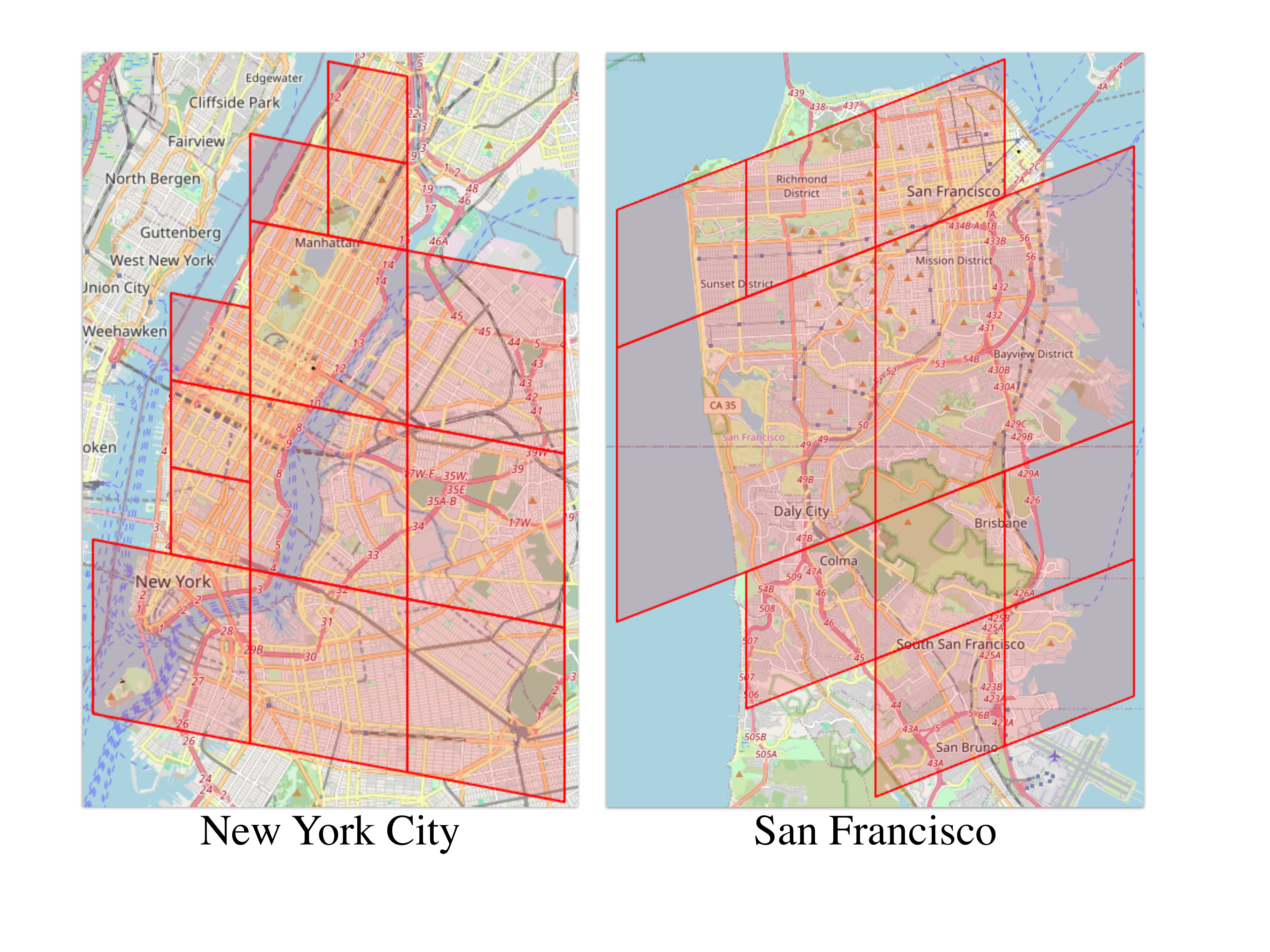}
\caption{
\textbf{Data Coverage.}
We show the region we sample our locations from in both cities in red.
Note that our data collection does not coverage the whole region of administration for both cities.
}
\label{fig:data_coverage}
\end{figure}
We collect our data from two major US cities, San Francisco and New York City.
In particular, we sample locations from a selected region of each city, as shown in \Fig{data_coverage}.
Note that the regions are selected to represent busy, dense regions of the city for our study of trends.
They do not cover the whole city by the definition of range of city administration.
Yet, within the selected regions we already collect 20M images for our massive scale study.

\subsection{Local Change Detection Details}
\label{sec:change_det_details}
\paragraph{Data Preparation.}
For the first step, local change detection, in Visual Chronicles, we first sample a set of locations from the city of interest.
Since the street view imagery we use is sampled from capture vehicles, we naturally have all the captured locations scattered on the street instead of inside of buildings.
We hence directly sample a set of locations from the capture locations.
To avoid redundancy caused by sampling nearby points, we implement Non-Maximum Suppression (NMS).
Specifically, we first calculate the number of nearby images (as defined in \Sec{local_change}) for each location.
Then we rank all locations by number of neighbors in deceasing oreder, and apply NMS to select non-redundant locations.
After selecting all locations, we collect local neighboring images following the process outlined in \Sec{local_change}.
Note that we filter out locations that have fewer than 10 local images at the end of this preparation step.

\begin{figure*}[t]
\noindent\makebox[\textwidth][c]{
\framebox{%
\begin{minipage}{0.98\linewidth}
\footnotesize
\textbf{\textit{Prompt:}}
Below are pictures of the same location taken over time.\\
\begin{center}
\begin{tabular}{cccc}
  &This is image No. 2, taken in June 2021. & This is image No. 3, taken in May 2022. & \\
  \textbf{...} &\includegraphics[width=0.3\textwidth]{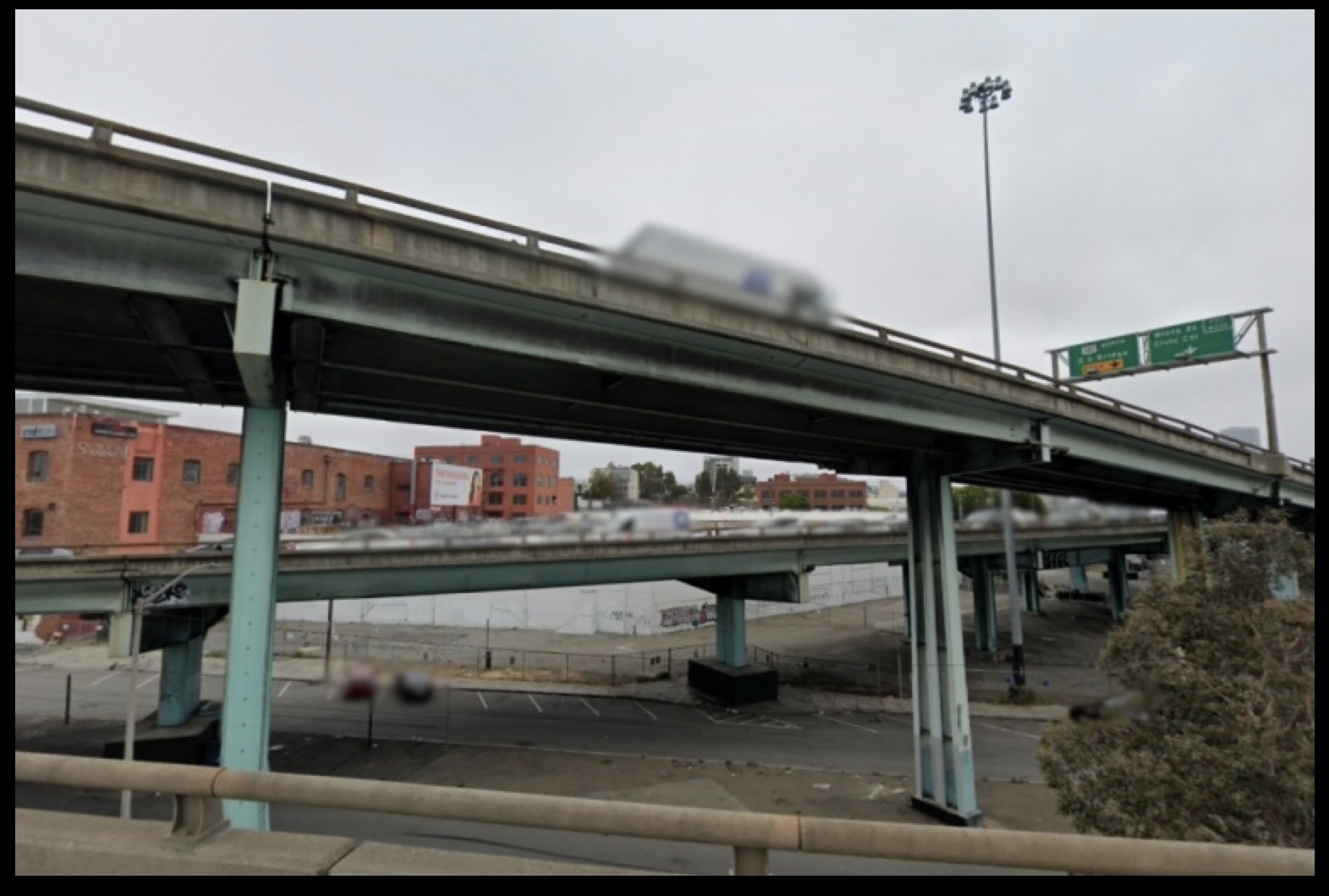} &
  \includegraphics[width=0.3\textwidth]{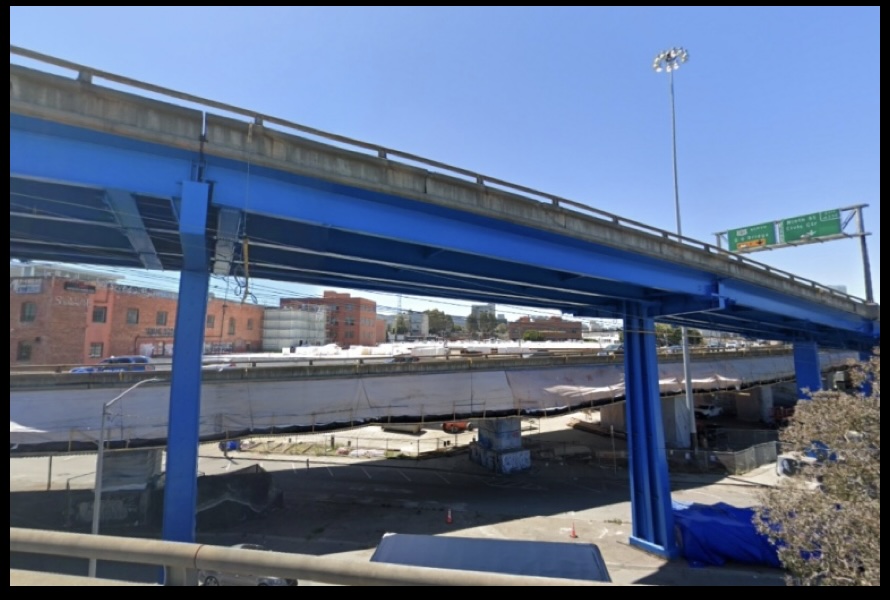}&\textbf{...}
\end{tabular}
\end{center}
You, as a brilliant {role}, need to describe what have changed over time in these images in details and also tell me from what images you found the changes.\\
Please format the answer strictly in the following format, one change a line:\\
Start: [Start\_DESCRIPTION] $\rightarrow$ End: [End\_DESCRIPTION] (happened after image No.[X]).\\
Format Instructions:\\
Replace [Start\_DESCRIPTION] with detailed descriptions of the start of the change;\\
Replace [End\_DESCRIPTION] with detailed descriptions of the end of the change;\\
Replace [X] with the index of the image where the change happened after. That said, this number should come from the "image No.[X]" descriptions above.\\

There are a few guidelines that you should follow when answering the question:
\begin{itemize}
\item DO NOT HALLUCINATE. You should stick to the changes that can be clearly and certainly seen in the images.
\item IGNORE CHANGES IN THE PERIPHERAL. Because the viewpoint may wiggle, ignore the changes that happened very close the the peripheral of images because they usually can not be confirmed.
\item IGNORE TINY UNRECOGNIZABLE TEXT. If certain text in the image is too small to be clearly recognized, such text content should not be part of a change because it can not be confirmed.
\item IGNORE CHANGES CAUSED BY OCCULUSIONS. There will be occlusions, like trucks or cars parked in front of some subjects in images. If a subject appear or disappear due to such occlusions, it is not a change.
\item TRACK THE SAME SUBJECT. When detect changes, please be sure to describe the change happen to the same subject.
\item DO NOT MISTAKE FOREGROUND AND BACKGROUND. Do not mistake the foreground subject, such a street sign, with attachments to the background subject, like signs on the building behind the street sign.
\item IGNORE CHANGES CAUSED BY ILLUMINATION CHANGES. Ignore subtle color changes caused by illumination changes, like colors become darker because of shadows, you can get the clue of illumination from the time of day, i.e. the time after ``taken at '' in the prompt.
\item ONE EVENT A CHANGE. For each detected change, it should focus on one event only, not multiple events. For example, the change ``Start: The storefront of the building has a black awning and a red truck parked in front of it. $\rightarrow$ End: The storefront of the building has a red awining now and the truck is gone'' should be 2 changes one on the awning and the other on the truck.
\item CHANGES SHOULD BE SIGNIFICANT. This means the description after ``Start:'' should describe a clearly different status of THE SAME SUBJECT from the description after ``End:''. For example, the change ``Start: The building is free of scaffolding. $\rightarrow$ End: There is no scaffolding on the building.'' is not a significant change because both Start and End are the same status of the same subject.
\end{itemize}

Now, please tell me changes that you can find in the images. Make sure you tell me ALL THE CORRECT CHANGES, following the above guidelines.\\

\textbf{\textit{MLLM (Gemini 1.5) Answers:}}\\

…\\
Start: The support beams of the overpass were in pale green color. $\rightarrow$ End: The support beams were painted into a bright blue color. (Happened after image No. 2)\\
…
\end{minipage}
}}
\caption{
\textbf{Local Change Detection Prompt}.
}
\label{fig:change_det_prompt}
\end{figure*}

\paragraph{Detection Prompting.}
To detect local changes using MLLMs, we essentially feed a sequence of images, in chronological order, to the MLLM and ask it 
what changes it detects and from which images.
We use Gemini-1.5 Pro (Gemini) as the MLLM here.
We find that such prompts already work surprisingly well on Gemini.
Gemini would describe significant temporal semantic changes it notices, including changes minute in visual scale such as the tiny security camera in \Fig{totd_nyc}.
At the same time, it often voluntarily ignores semantically insignificant changes such as illumination changes or camera displacement changes (as our camera poses are not perfectly aligned).
Yet, through our empirical study, we find a few error patterns of such simple prompting.
Among them, the notable ones include:
1) Gemini may hallucinate changes that can not be certainly seen in images.
2) The answers occasionally include changes that are caused by camera wiggles, normally in the periphery of the frame.
3) Sometimes Gemini would mistake color changes caused by illumination change as a permanent change, \eg change from a dark color in a shadow to a very bright color under the sun.
4) Very rarely, Gemini may mistake the foreground object with the background, \eg mistake the sign and text on the car in front of a building as text and sign on the building.
To address these, we articulate a more extensive list of instructions in the prompt shown in \Fig{change_det_prompt}.
We include all the error patterns we empirically find in this prompt.

\paragraph{Self-Critic.}
Despite the careful prompting, we still find Gemini making mistakes such as hallucination in some of the answers.
Inspired by OpenAI o1, we implement a simple Self-Critic strategy by feeding the visual evidence and the detected change back to Gemini and ask it to verify that the change is significant and that it can be found in the visual evidence. We find such a simple verification step greatly improves the precision of local change detection, as shown in \Sec{change_det_ablation}.

\subsection{Trend Discovery Details}
\label{sec:trend_discov_details}
\paragraph{Trend Proposal.}
In the second step of our system, we derive trend proposals from detected local changes.
This step has $2$ parts.
First, we design a prompt that make MLLMs deduct potential proposals from a detected change, as shown in \Fig{trend_prop_prompt}.
Then, we implement a standard Canopy Clustering~\cite{mccallum2000efficient} using text embedding distance and set the tight distance threshold to $0.15$ and the loss distance threshold to $0.2$.
Because our verification needs at least $k$ nearest neighbours in cluster, we filter clusters from the Canopy Clustering that has less than $k$ elements.
After the filtering, each cluster center becomes a trend proposal, and all its derived trends text will be candidates for trend verification.

\paragraph{Verification Prompting.}
We develop a detailed prompt promoting MLLM (in our system, Gemini) to verify if a change belongs to a trend proposal, shown in \Fig{trend_verify_prompt}.
In particular, we leverage the in-context learning ability of MLLMs, by including a small set of examples in the prompt.
We empirically find that such examples improve the verification quality.

\subsection{Evaluation Details}
\label{sec:eval_details}
For our evaluation in \Sec{change_det_eval} and \Sec{trend_discov_eval}, we sample the test set from our data collection from New York City.
For the evaluation of change detection in \Tab{change_det}, we randomly sample $200$ locations in New York City and ask a human labeller to describe a list of changes for each location and from which $2$ consecutive images the changes were found, as the ground truth changes.
Given such labels, we can derive if there's a change between any $2$ consecutive images, which is the ground truth we use to calculate the Average Precision in \Tab{change_det}.
In our evaluation of trend discovery in \Tab{trend_discov}, we aim to collect $50$ clusters each with $40$ candidate changes to verify.
However, na\"ively randomly sampling such a collection would lead to mostly negative labels, that is, the change does not belong to the trend.
To better balance the test labels, we over-sample $75$ clusters, with a selection heuristic of enough nearby neighbours, \ie more than $N$ changes in the pool with text embedding distance smaller than $0.25$. Thereafter, to sample balanced change candidates, we over-label change changes until we collect at least $20$ positive labels for the cluster, \ie $50\%$ positive rate. We also apply early-stop, \ie setting the maximum number of changes to label for each cluster, to avoid excessive labelling for one cluster. After labelling all $75$ clusters, we pick $50$ clusters with the most balanced test labels as our final test set.
For \Tab{hybrid}, since NV-Emb is predicting continuous distance values, we grid-search $1000$ thresholds to binarize its predictions and pick the highest accuracy to report.

\section{Extended Evaluations}
\label{sec:more_eval}
\subsection{Validity of Evaluation Data}
\label{sec:eval_validity}
Our analysis is conducted on massive-scale datasets.
It's impractical to annotate even $1\%$ of the $20$M images in the datasets.
We instead designed an evaluation set large enough to provide comparisons of different methods with low variance, with conclusions generalizable to larger-scale applications.
To verify this, we computed the standard deviation of metrics using $1$K random $75\%$-subsets of the evaluation set, and found them to be $\sim1\%$ in AP for both subtasks.

\subsection{Ablation Studies}
\label{sec:all_ablations}
We conducted extensive ablation studies evaluating design choices of our system, including MLLMs for change detection (\Sec{change_det_eval} \& \Tab{change_det}), prompting strategy in change detection (\Sec{change_det_ablation} \& \Tab{ablation_change_det}), MLLMs for trend discovery (\Sec{trend_discov_eval} \& \Tab{trend_discov}), hybrid solutions for scaling (\Sec{trend_discov_eval} \& \Tab{hybrid}), its hyperparameters (\Sec{trend_discov_ablation} \& \Tab{ablation_hybrid}), and alternative MLLMs (\Sec{mllm_ablation} \& \Tab{mllm_ablation}).

\subsection{Alternative MLLMs}
\label{sec:mllm_ablation}
\begin{table}
\centering{
\resizebox{0.83\columnwidth}{!}{
\begin{tabular}{lcc}
\toprule
MLLM & Change Detection & Trend Discovery \\
\midrule
LLAVA~\cite{liu2023improvedllava} & $64.31 \%$ & $81.11 \%$ \\
Gemini~\cite{reid2024gemini} & $76.56 \%$ & $86.63 \%$ \\
\bottomrule
\end{tabular}
} 
} 
\caption{\textbf{Ablation on Different MLLMs} in AP.
The smaller MLLM, LLAVA, still performs decently on both sub-problems, while being slightly worse than more powerful Gemini.
}
\label{tbl:mllm_ablation}
\end{table}
We investigate the impact of using different MLLMs on the effectiveness of our system.
To this end, we replace Gemini with a smaller open-sourced MLLM, LLAVA~\cite{liu2023improvedllava}, and evaluate its performance on both sub-problems in our system.
We find that it still produces reasonably accurate analysis while being slightly worse than Gemini.

\subsection{Self-Critic Prompting in Change Detection}
\label{sec:change_det_ablation}
\begin{table}
\centering{
\resizebox{0.8\columnwidth}{!}{
\begin{tabular}{lcc}
\toprule
 & Without Self-Critic & With Self-Critic \\
\midrule
\textbf{Precision} & $57.83 \%$ & $81.74 \%$ \\
\textbf{Recall} & $91.50 \%$ & $89.87 \%$ \\
\bottomrule
\end{tabular}
} 
} 
\caption{\textbf{Ablation on Self-Critic in Local Change Detection}.
The Self-Critic strategy (\Sec{change_det_details}) significantly improves the precision of local change detection.
}
\label{tbl:ablation_change_det}
\end{table}
As mentioned in \Sec{change_det_details}, we find that applying a round of self-critic to local change detection greatly improves the precision of the detection.  In \Tab{ablation_change_det}, we quantitatively evaluate the impact of self-critic.
We find that it boosts the precision by a significant margin of more than $33\%$ while not degrading much on the recall.

\subsection{Hyperparameters in Trend Discovery}
\label{sec:trend_discov_ablation}
\begin{table}
\centering{
\resizebox{.98\columnwidth}{!}{
\begin{tabular}{lcccc}
\toprule
\textbf{Method} & $k=2N$ & $k=3N$ & $k=4N$ & $k=5N$ \\
\midrule
\textbf{Acc.@$50$} & $88.6\%$ & $93.9\%$ & $95.9\%$ & $97.2\%$ \\
\textbf{Acc.@$100$} & $91.4\%$ & $94.6\%$ & $96.1\%$ & $97.1\%$ \\
\textbf{Acc.@$200$} & $95.7\%$ & $98.3\%$ & $99.2\%$ & $99.5\%$ \\
\bottomrule
\end{tabular}
} 
}
\caption{\textbf{Ablation of Hyperparameters of Our Hybrid Solution in Trend Discovery}.
$k$ is the number of nearest neighbours our hybrid solution verify using MLLMs. $N$ is the frequency threshold of a trend.
We choose $k=3N$ for all our experiments for its balance between accuracy and efficiency.
}
\label{tbl:ablation_hybrid}
\end{table}
Our hybrid solution to trend discovery requires setting a hyper-parameter $k$, \ie the number of nearest changes for which we use MLLMs to verify regarding a trend proposal.
In \Tab{ablation_hybrid}, we show an ablation study of varying the value of $k$, as multiples of the trend threshold $N$.
We find that as $k$ increases, the accuracy goes up as expected, which comes at a cost of more MLLM inferences and slower verification.
However, $k=3N$ can already produce satisfying accuracy ($>90\%$ across the board).
Hence, we choose it as a good balance between accuracy and speed.

\subsection{MLLMs Without Images}
\label{sec:no_images}
\paragraph{Complete Answers from MLLMs.}
As mentioned in \Sec{mllm_compare}, we find MLLMs' answers without looking at any images unsatisfactory.
Such answers are often abstract and lack intriguing details.
In \Fig{mllm_wo_images}, we illustrate the complete answers from MLLMs to the prompting questions, ``What are the trends of changes happened in [NYC / SF] that we can see visually in 2011-2023?''.
Here we use Gemini-1.5 Pro for queries for both cities.

\begin{figure*}[t]
\noindent\makebox[\textwidth][c]{
\framebox{%
\begin{minipage}{0.98\linewidth}
\footnotesize
\textbf{\textit{Prompt:}}
What are the trends of changes happened in {\textbf{New York City}} that we can see visually in 2011-2023?\\

\textbf{\textit{MLLM (GPT-4o) Answers:}}\\
\begin{tabular}{cccc}
  \includegraphics[width=0.22\textwidth]{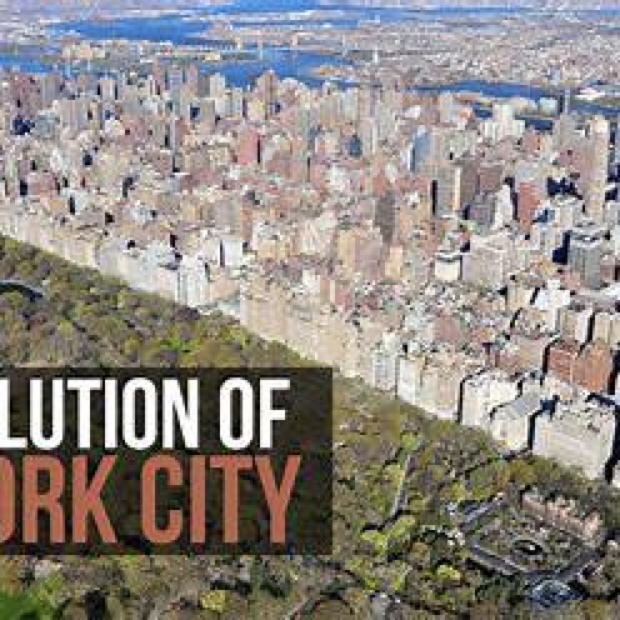} &
  \includegraphics[width=0.22\textwidth]{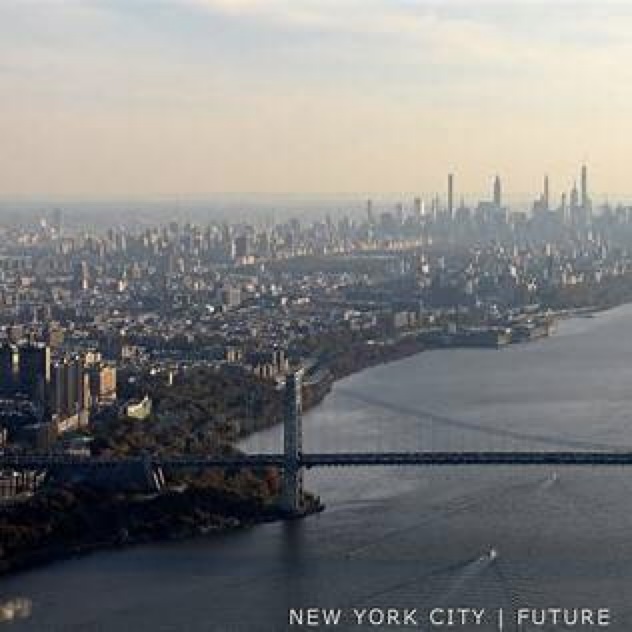} &
  \includegraphics[width=0.22\textwidth]{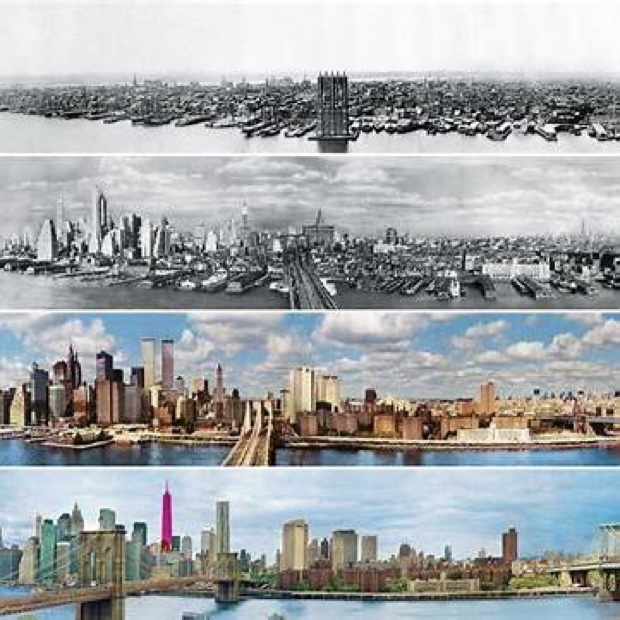} &
  \includegraphics[width=0.22\textwidth]{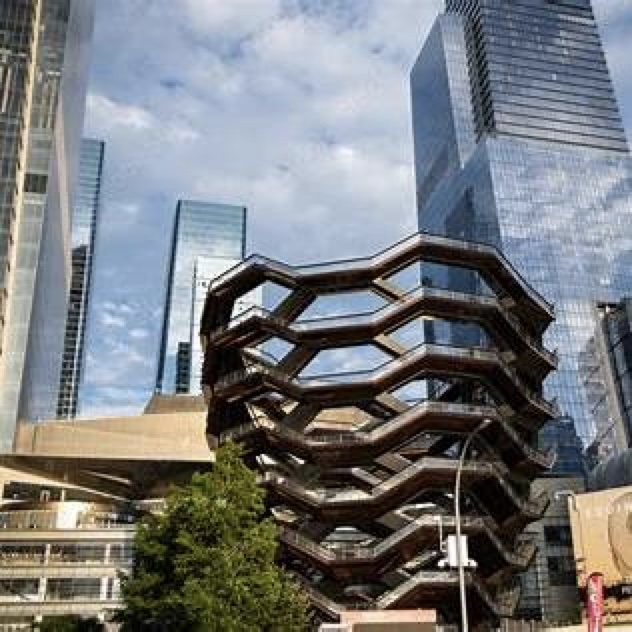} 
\end{tabular}
Between 2011 and 2023, New York City underwent significant visual transformations across its skyline, infrastructure, and public spaces. Key trends include:
\begin{enumerate}
    \item \textbf{Skyline Evolution:}
    \begin{itemize}
        \item \textit{Rise of Super-Tall Skyscrapers:} The city saw the construction of numerous super-tall buildings, particularly along 57th Street, known as “Billionaire’s Row.” Notable additions include 432 Park Avenue and Central Park Tower, which have redefined the city’s silhouette.
    \end{itemize}
    \item \textbf{Infrastructure and Public Space Enhancements:}
    \begin{itemize}
        \item \textit{Pedestrianization Initiatives:} Broadway underwent significant changes, with sections transformed into pedestrian plazas to enhance walkability and reduce vehicular traffic. Notably, areas in Times Square and Herald Square were closed to cars, creating more space for pedestrians.
        \textit{Reopening of Public Spaces:} After 15 years, a 15,000-square-foot greenspace under the Brooklyn Bridge, part of “The Arches” area, reopened in November 2024, providing much-needed greenery and recreational space for the community.
    \end{itemize}
    \item \textbf{Adaptive Reuse of Buildings:}
    \begin{itemize}
        \item \textit{Office-to-Residential Conversions:} In response to changing work habits and increased office vacancies, developers have been converting obsolete office buildings into residential apartments, particularly in areas like the Financial District. This trend aims to address housing shortages and revitalize underused spaces.
    \end{itemize}

    \item \textbf{Preservation and Restoration Efforts:}
    \begin{itemize}
    \item \textit{Neon Signage Conservation:} Efforts to preserve iconic neon signs have intensified, with restorations of landmarks like Radio City Music Hall and the Essex House hotel, reflecting a desire to maintain the city’s historical visual elements amid modernization.
    \end{itemize}
    \item \textbf{Sustainable Urban Design:}
    \begin{itemize}
        \item \textit{Green Infrastructure Projects:} The city has embraced sustainable development, with projects like the transformation of Fifth Avenue into a more pedestrian-friendly boulevard, featuring expanded sidewalks, seating, and greenery, aiming to enhance urban vitality and environmental sustainability.
    \end{itemize}
\end{enumerate}
These developments collectively highlight New York City’s dynamic evolution, balancing modernization with preservation and sustainability.
\end{minipage}
}}
\caption{
\textbf{GPT-4o Predictions Without Images. (Accessed time: {Nov. 2024})}
}
\label{fig:alt_mllms}
\end{figure*}

\paragraph{Alternative Prompts and MLLMs.}
We also vary the wording of the prompts in search for potentially better alternatives,
such as replacing ``changes ... that we can see visually'' with ``visual changes'' or simply ``changes''.
Yet, we empirically find the results not varying much in quality, except we see more abstract trends, \eg ``Increased focus on sustainability'', when using simply ``changes''.
Additionally, we show in \Fig{alt_mllms} the results from an alternative MLLM, namely GPT-4o.
We find that GPT-4o focuses more on unique changes, such as ``construction of numerous super-tall buildings ... along 57th Street ... known as Billionaire’s Row.'', but misses more frequent trending changes such as bike lane expansions, which both Gemini and Visual Chronicles identify as a significant trend.
We also notice that to answer the query, GPT-4o would search online and retrieve relevant information, both image and text, in the answering process, and even output a few images in the output, as shown in \Fig{alt_mllms}. Yet these images are not strictly all visual evidence for the discovered trends, unlike in Visual Chronicles.

\subsection{MLLMs Processing All Images}
\label{sec:mllm_all_images}
In \Sec{mllm_compare}, we discussed the context size as MLLMs' major limitation.
Due to such constraint, it is infeasible to feed all our images to MLLMs.
Instead, in \Fig{mllm_all_images}, we sub-sample the maximum subset of our images that can be fed into MLLMs, ranging from 7K to 8K images because we feed complete image sequences to MLLMs and such sequences vary in lengths.
Specifically, we sample $4$ such subsets and take the union of the answers as the final answer shown in \Fig{mllm_all_images}.
Note that we only sample $4$ subsets for increasing the number of subsets mainly creates duplicate trends.

\subsection{MLLMs Processing All Changes}
\label{sec:mllm_all_changes}
As mentioned in \Sec{trend_discov_eval}, trend discovery is not trivial to MLLMs for simply feeding all change descriptions to MLLMs is not adequate.
In \Fig{mllm_all_changes}, we show the complete results of feeding as many changes as possible to MLLMs and ask for trends among them.
Here we sample $50$K changes for each batch to feed to MLLMs, from our pool of $3$M changes.
We sample $4$ batches in total and take the union of answers from each batch as the final answer shown in \Fig{mllm_all_changes}.
Note that we only sample $4$ batches because in our observations, more batches mostly create redundancy in found trends.

\subsection{Hybrid Trend Discovery Method and RAG}
\label{sec:connection_to_rag}
Our hybrid solution to trend discovery bears some connections to Retrieval-Augmented-Generation (RAG~\cite{lewis2020retrieval}).
Particularly, both use embeddings to filter relevant examples from a database.
However, ours is fundamental different from RAG in a few aspects:
1) Our goal is different from RAG. We aim to find salient patterns of examples in the database, whereas RAG retrieves examples to assist question answering.
2) The scale of relevant examples is different. In our task, we mark thousands of examples in the database as relevant, while RAG normally select several most relevant examples.
3) We use relevant examples differently from RAG. RAG often combines all relevant examples together as context to LLM or MLLMs. In contrast, we process each relevant example independently to verify visual trend proposals.

\section{Limitations and Failure Cases}
\label{sec:failures}
As described in the Conclusion section of the main paper, our approach has limitations that lead to failure cases and topics for future work.   The following provides a few examples.

The proposed approach is natural when the input image collection can be decomposed into small sets for MLLM analysis.  We utilized image poses to create sets of images at nearby viewpoints from which local changes can be detected.  However, this method would not work well for detecting all types of trends, including ones that involve global changes or ones that cannot be observed in images.

We rely upon simple prompts to guide the MLLM to produce descriptions of local changes that we expect people would find interesting.  However, other types of changes might be of interest in different applications (e.g., to find new dirt on the sidewalk).  That would require prompt engineering to get the MLLM to detect certain types of changes -- further study is required to discover how readily good prompts can be found.

Our formulation produces sets of visual changes that happened multiple times at different locations, but does not reason about how often the reverse change happened, and so it sometimes reports changes that are the result of transient activities (e.g., the addition of Halloween decorations), which are not in fact long-term trends.   Similarly, our system does not measure the fraction of times a change happened for the same starting state, which could be valuable for studying certain types of trends.   Adding these capabilities is an interesting topic for future studies.

\section{Potential Societal Impacts}
\label{sec:social_impact}
To our best knowledge, Visual Chronicles marks the first proof-of-concept of using MLLMs to analyze millions of images.
In the context of searching for trends of visual changes in the urban environment, our studies show that a system like ours could potentially facilitate studies in urban planning, civil engineering, or public policy, by automatically providing insights from massive scale dataset.
Nevertheless, as we mentioned in our discussion on our results, it's vital to keep aware of the potential bias in our data, hence being cautious in drawing conclusions that demand statistical soundness. 
Particularly, careful uses of systems of this kind are critical in battling misinformation.
Our focus on providing verifiable visual evidence for any trend proposed by our system can be seen as part of such efforts.

\begin{figure*}[t]
\noindent\makebox[\textwidth][c]{
\begin{minipage}{1.0\linewidth}
\framebox{%
\begin{minipage}{0.98\linewidth}
\footnotesize
\textbf{\textit{Prompt:}}
What are the trends of changes happened in {\textbf{New York City}} that we can see visually in 2011-2023?\\

\textbf{\textit{MLLM (Gemini 1.5) Answers:}}
Visually noticeable changes in New York City between 2011 and 2023 include:

\textbf{Infrastructure \& Development:}
\begin{itemize}
    \item \textbf{Rise of Supertalls:} The proliferation of extremely tall, slender skyscrapers, particularly in Midtown Manhattan (e.g., 432 Park Avenue, 111 West 57th Street) and Downtown Brooklyn, dramatically altered the skyline.
    \item \textbf{Hudson Yards Development:} The massive Hudson Yards project on Manhattan’s west side, with its distinctive Vessel structure and surrounding high-rises, became a completely new neighborhood, visually impacting the cityscape.
    \item \textbf{Second Avenue Subway:} The long-awaited Second Avenue Subway extension finally opened its first phase, visibly changing the streetscape along Second Avenue on the Upper East Side with new station entrances.
    \item \textbf{Brooklyn Waterfront Transformation:} Continued development along the Brooklyn waterfront, from Williamsburg to DUMBO to Greenpoint, resulted in new parks, residential buildings, and commercial spaces, reshaping the view from Manhattan.
    \item \textbf{Bike Lane Expansion:} A significant increase in dedicated bike lanes throughout the city is a visible change, altering street markings and traffic patterns.
\end{itemize}
\textbf{Social \& Cultural:}
\begin{itemize}
    \item \textbf{Increased Outdoor Dining:} The COVID-19 pandemic led to a boom in outdoor dining structures, from simple sheds to elaborate setups, significantly changing the look and feel of many streets. While some have been removed, many remain and have become a more permanent fixture.
    \item \textbf{Public Art Installations:} Larger and more frequent public art installations have become more common in parks and public spaces, adding visual interest to the city.
    \item \textbf{Retail Landscape Changes:} The rise of online shopping and the pandemic impacted brick-and-mortar stores, resulting in visible vacancies and changes in storefront usage in some areas. 
    \item \textbf{Greening Initiatives:} Efforts to increase green space, including park expansions and rooftop gardens, are slowly but surely adding more visible greenery to the city.
\end{itemize}

\textbf{Less Visible but Noteworthy:}
\begin{itemize}
    \item \textbf{Demographic Shifts:} While not immediately visually obvious, changes in demographics (gentrification, population growth/decline in certain neighborhoods) influence the types of businesses, public services, and overall aesthetic of different areas over time.
    \item \textbf{Security Measures:} Increased security measures after 9/11 and other events, though not always visually prominent, have subtly altered public spaces and infrastructure.
\end{itemize}
To truly appreciate these changes, comparing photos of specific locations from 2011 and 2023 can be particularly revealing. Google Street View’s historical imagery feature is a useful tool for this purpose. You’ll see the transformation of specific streetscapes and neighborhoods unfold before your eyes.
\end{minipage}
}\vspace{0.5em}
\framebox{%
\begin{minipage}{0.98\linewidth}
\footnotesize
\textbf{\textit{Prompt:}}
What are the trends of changes happened in {\textbf{San Francisco}} that we can see visually in 2011-2023?\\

\textbf{\textit{MLLM (Gemini 1.5) Answers:}}
Visually noticeable changes in San Francisco between 2011 and 2023 include:

\textbf{Increased Construction and Development:}
\begin{itemize}
    \item \textbf{High-rise buildings:} The skyline has grown denser, particularly in areas like South of Market (SoMa), Mission Bay, and Transbay. Many new residential and commercial high-rises have been constructed.
    \item \textbf{Crane activity:} While fluctuating, crane activity has been generally high throughout this period, a visible sign of ongoing construction.
    \item \textbf{Infilling:} Vacant lots and older, smaller buildings have been replaced by larger, denser structures in many neighborhoods.
\end{itemize}
\textbf{Transportation Changes:}
\begin{itemize}
    \item \textbf{Bike lanes and infrastructure:} A significant increase in protected and dedicated bike lanes is visible throughout the city, reflecting the push for alternative transportation.
    \item \textbf{Transit changes:} While the Central Subway finally opened in this period, other visual changes are less noticeable from street level. However, increasing numbers of electric buses might be seen.
    \item \textbf{Scooter proliferation (and subsequent regulation):} From around 2017-2019, dockless scooters flooded the city, dramatically changing the streetscape, before regulations brought more order.
\end{itemize}
\textbf{Neighborhood Transformations:}
\begin{itemize}
    \item \textbf{Mid-Market revitalization (and struggles):} The area around Twitter’s headquarters has seen noticeable changes, with new businesses and offices, but also persistent challenges with homelessness and public safety.
    \item \textbf{Mission Bay development:} This formerly industrial area has transformed into a bustling neighborhood with new housing, UCSF facilities, and the Chase Center.
    \item \textbf{Gentrification and displacement:} While not always directly visible, the effects of gentrification can manifest in changes to storefronts, the types of businesses present, and the overall aesthetic of a neighborhood.
\end{itemize}
\textbf{Other Visible Changes:}
\begin{itemize}
    \item \textbf{Increased homelessness:} Sadly, a significant rise in visible homelessness has been a major change in this period, with tent encampments and individuals on the streets becoming more prevalent.
    \item \textbf{Retail vacancies:} Particularly after the pandemic, an increase in empty storefronts, especially in downtown areas, has been a noticeable change.
    \item \textbf{Parklets and outdoor dining:} Accelerated by the pandemic, parklets and outdoor dining spaces became more common, altering the look and feel of many commercial streets. 
    \item \textbf{Art installations and murals:} San Francisco’s vibrant street art scene continues to evolve, with new murals and installations regularly appearing and disappearing.
\end{itemize}
To actually see these changes, you would need to compare imagery from the two periods:
\begin{itemize}
    \item \textbf{Google Street View:} Offers a historical record of street-level views, allowing for direct comparisons over time.
    \item \textbf{Satellite imagery:} Services like Google Earth can show broader changes in building density and land use. 
    \item \textbf{Photo archives:} Local historical societies, libraries, and online photo collections can provide visual documentation of past cityscapes.
    \item \textbf{Documentaries and news footage:} Videos can offer a dynamic perspective on how the city has changed over time.
\end{itemize}
By comparing imagery and exploring these different sources, you can gain a clearer visual understanding of San Francisco’s transformation between 2011 and 2023.
\end{minipage}
}
\end{minipage}}
\caption{
\textbf{Full Gemini 1.5 Predictions Without Images  (Accessed time: {Nov. 2024}).}
}
\label{fig:mllm_wo_images}
\end{figure*}
\begin{figure*}[t]
\noindent\makebox[\textwidth][c]{
\framebox{%
\begin{minipage}{0.98\linewidth}
\footnotesize
\textbf{\textit{Prompt:}}
I have some image sequences, each captured from one location over time. They are sorted in chronological order.

\begin{center}
    [… Image Sequences Containing 7K-8K Images…]
\end{center}

Now you, as a brilliant analyst, need to tell me what are the trends of temporal changes in these sequences. Describe each trend in a line in the format of ``[Before] $\rightarrow$ [After] (Evidence ID: [IDX],[IDX])''. 

Replace [Before] with descriptions for before the trend, replace [After] with descriptions for after the trend, and replace [IDX] with an evidence image sequence index for the trend. For each trend, you need to tell me at least 500 evidence.\\

\textbf{\textit{MLLM (Gemini 1.5) Answers:}}
\begin{itemize}
    \item Construction site with a partially built structure $\rightarrow$ Completed building.
    \item One business $\rightarrow$ Different business
    \item One business $\rightarrow$ Vacancy/Lease available
    \item Older signage/awning $\rightarrow$ Newer signage/awning
    No bike lane markings $\rightarrow$ Bike lane markings added
    \item One advertisement $\rightarrow$ Different advertisement
    \item One business storefront $\rightarrow$ Different business storefront
    \item Older storefronts/signage $\rightarrow$ New storefronts/signage
    Bare trees/winter scenery $\rightarrow$ Leafy trees/summer scenery
    \item Building under scaffolding $\rightarrow$ Building with no scaffolding
    \item Brownstone building at a street corner $\rightarrow$ Brownstone building with an awning and signage
    \item One storefront/business signage $\rightarrow$ Different storefront/business signage/closed down
    \item Smaller plants/Bare trees $\rightarrow$ Larger plants/Leafy trees/Bare Trees
    \item Older graffiti $\rightarrow$ New graffiti/graffiti removal
    \item Building exterior or area without scaffolding/construction equipment $\rightarrow$ Building exterior or area with scaffolding/construction equipment
\end{itemize}
\end{minipage}
}}
\caption{
\textbf{MLLMs Processing All Images.}
}
\label{fig:mllm_all_images}
\end{figure*} 
\begin{figure*}[t]
\noindent\makebox[\textwidth][c]{
\framebox{%
\begin{minipage}{0.98\linewidth}
\footnotesize
\textbf{\textit{Prompt:}}
Here are 50K text descriptions, each describing a change happened in the city.

\begin{center}
    [… Contents of 50K Changes …]
\end{center}

Now you, as a brilliant analyst, need to tell me what are the trends among these changes. Describe each trend in a line in the format of ``[Before] $\rightarrow$ [After] (Evidence ID: [IDX],[IDX])''.

Replace [Before] with descriptions for before the trend, replace [After] with descriptions for after the trend, and replace [IDX] with an evidence change index for the trend. For each trend, you need to tell me at least 500 evidence.\\

\textbf{\textit{MLLM (Gemini 1.5) Answers:}}
\begin{itemize}
    \item There is no scaffolding on the building $\rightarrow$ There is scaffolding on the building.
    \item The scaffolding is present.  $\rightarrow$ The scaffolding is gone.
    \item The crosswalk markings are solid white lines  $\rightarrow$ The crosswalk markings are thick white stripes.
    \item The building is covered in scaffolding  $\rightarrow$ The scaffolding is removed.
    \item The building has a [Business\_name] sign  $\rightarrow$ The [Business\_name] sign is replaced.
    \item The storefront is occupied by a business  $\rightarrow$ The storefront is vacant.
    \item The storefront is occupied by one business  $\rightarrow$ The storefront is occupied by another business.
    \item The trees on the street is thin  $\rightarrow$ The trees grow taller and thicker.
    \item Scaffolding is erected in front of the building  $\rightarrow$ The scaffolding is removed, revealing the building's facade.
    \item The billboard has an advertisement  $\rightarrow$ The advertisement changed.
    \item A business with [Business\_name] sign  $\rightarrow$ The [Business\_name] sign is gone.
    \item There were no bike lanes  $\rightarrow$ There is a new bike lane.
    \item Entrance to the building has a [Color] door  $\rightarrow$ The entrance to the building has a [Color] door.
    \item Green scaffolding covers the building  $\rightarrow$ The scaffolding is gone.
\end{itemize}
\end{minipage}
}}
\caption{
\textbf{MLLMs Processing All Changes.}
}
\label{fig:mllm_all_changes}
\end{figure*} 
\begin{figure*}[t]
\noindent\makebox[\textwidth][c]{
\framebox{%
\begin{minipage}{0.98\linewidth}
\footnotesize
\textbf{\textit{Prompt:}}

I have a text description of a change that happened in a city. For example:
Before: The Ginger Man's Bar did not have any ``Bar for Lease'' sign.\\
After: The Ginger Man's Bar put up a ``Bar for Lease'' sign on the storefront window.\\
As you can see, in the description, there will be the status before the change and the status after the change.\\
Your job, as a brilliant analyst, is to derive more abstract versions of the same change by removing details.\\

Formulate the answer strictly in the following list format:\\
There are [N] levels of details on where the change happened:\\
p[X]. [Description of one level]\\
Meanwhile, there are [M] levels of details on the change itself:\\
c[X]. [Description of one level]\\

[More descriptions of the reasoning].\\

Answer:
(reason) [A derived abstract version].\\

Here are a few examples:

… [Examples] …\\

Here is the change you need to derive abstract versions for:\\
Before: The single door to the right of the main entrance has a sign above it that reads ``151''.\\
After: The sign above the single door is blank.\\

Now, please tell me the abstract versions of this change, following the above instructions.\\

\textbf{\textit{MLLM (Gemini 1.5) Answers:}}

Derivation:\\
There are 3 levels of details on where the change happened:
\begin{itemize}
    \item p1. The single door to the right of the main entrance. (reason: the original very specific place).
    \item p2. The single door. (reason: a more general location description without mentioning it's to the right of the main entrance).
    \item p3. A door. (reason: the most general location description).
Meanwhile there are 2 levels of details on the change itself:
    \item c1. The sign above it changing from reading ``151'' to being blank. (reason: the original very specific change.)
    \item c2. The sign above it changing from reading ``151'' to something else. (reason: more general without specifying the sign being blank.)
    \item c3. The sign above it changing. (reason: the most general change, without even specifying how the sign changed).
\end{itemize}

Answer:
\begin{itemize}
    \item (p1 + c1) The sign above the single door to the right of the main entrance changed from reading ``151'' to being blank.
    \item (p1 + c2) The sign above the single door to the right of the main entrance changed from reading ``151'' to something else.
    \item (p1 + c3) The sign above the single door to the right of the main entrance changed.
    \item (p2 + c1) The sign above the single door changed from reading ``151'' to being blank.
    \item (p2 + c2) The sign above the single door changed from reading ``151'' to something else.
    \item (p2 + c3) The sign above the single door changed.
    \item (p3 + c1) The sign above the door changed from reading ``151'' to being blank.
    \item (p3 + c2) The sign above the door changed from reading ``151'' to something else.
    \item (p3 + c3) The sign above the door changed.
\end{itemize}
\end{minipage}
}}
\caption{
\textbf{Trend Proposal Prompt}.
}
\label{fig:trend_prop_prompt}
\end{figure*} 
\begin{figure*}[t]
\noindent\makebox[\textwidth][c]{
\framebox{%
\begin{minipage}{0.98\linewidth}
\footnotesize
\textbf{\textit{Prompt:}}

I have a text description of a change that happened in a city. For example:
Before: The Ginger Man's Bar did not have any ``Bar for Lease'' sign.\\
After : The Ginger Man's Bar put up a "Bar for Lease" sign on the storefront window.\\
As you can see, in the description, there will be the status before the change and the status after the change.\\
I also have a test description of a group of similar changes that happened. For example:\\
A storefront put up a ``For Lease'' sign on its window.\\
Your job, as a brilliant analyst, is to decide if the given change belongs to the given group, ``Y'' for yes and "N" for no, and why.\\

Formulate the answer strictly in the following format:\\
Answer: [Y/N].\\
Reason: [Description of the reasoning].\\
For the example above, the answer would be:\\
Answer: Y.\\
Reason: The given change describes a bar putting up a ``Bar for Lease'' sign. We can remove the details of the location being a a specific "Ginger Man's Bar" but only keep it as a storefront, Meanwhile, we can keep the event of putting up the sign the same mostly intact, only removing the detail of a bar, i.e. ``Bar for Lease'' to ``For Lease'', to align with a storefront. Then, we can arrive at the group description. Therefore, the given change belongs to the given group.\\

Here are a few more examples:\\

… [Examples] …\\

Here is the change for you:\\
Before: The storefront on the corner is a GNC store.\\
After: The storefront is an OPT X store.\\

Here is the group for you:\\
A juice shop opened at the storefront.\\

\textbf{\textit{MLLM (Gemini 1.5) Answers:}}\\
Answer: N.\\
Reason: The group description specifies a change to a ``juice shop''. The provided change describes a change from a ``GNC store'' to an ``OPT X store''.  While both describe storefront changes, the specific type of store in the group description (``juice shop'') is not reflected in the change description.  We cannot generalize ``OPT X store'' to ``juice shop'' without adding information.
\end{minipage}
}}
\caption{
\textbf{Trend Verification Prompt}.
}
\label{fig:trend_verify_prompt}
\end{figure*} 

\end{document}